\documentclass[10pt,journal,compsoc]{IEEEtran}
 \pdfoutput=1   

\usepackage[labelfont=bf]{caption}
\usepackage[labelformat=simple]{subcaption}
\DeclareCaptionLabelSeparator{periodspace}{.\quad}
\usepackage{chngcntr} 
\usepackage{tabularx,amsmath,amssymb,amsthm,graphicx,float,wrapfig,multicol,epsfig,float,subcaption,amsfonts,amsbsy,amstext,mathrsfs,latexsym,color,amsthm,xcolor,fontenc,bm,mathtools,makecell,booktabs,doi,url}
\newcommand{\ra}[1]{\renewcommand{\arraystretch}{#1}}
\newcommand{\rulesep}{\unskip\ \vrule\ }
\usepackage[T1]{fontenc}

\graphicspath{{figures/}} 
\usepackage{subcaption}

\DeclareMathOperator*{\argmax}{argmax} 

\newcommand{\bSigma}{\bm{\Sigma}}
\newcommand{\bx}{\mathbf{x}}
\newcommand{\by}{\mathbf{y}}
\newcommand{\bz}{\mathbf{z}}
\newcommand{\D}{\mathbf{D}}
\newcommand{\bxi}{\boldsymbol\xi}
\newcommand{\R}{\mathbb R}

\definecolor{NUSBlue}{RGB}{0,61,124} 
\definecolor{NUSOrange}{RGB}{239,124,0}

\usepackage{cleveref}
\AtBeginDocument{%
}

\definecolor{lgray}{gray}{0.6}

\usepackage{hyperref}
\hypersetup{
  colorlinks= true, 		 
  urlcolor  = NUSBlue, 		 
  linkcolor = NUSBlue, 	     
  citecolor = NUSBlue} 	 	 

  \usepackage{cite}
\ifCLASSINFOpdf
\else
\fi
\usepackage{amsmath}

\begin{document}
%
\title{Unsupervised Statistical Learning for \\ Die Analysis in Ancient Numismatics}
%
%
%

\author{Andreas~Heinecke, 
        Emanuel~Mayer, Abhinav~Natarajan, and~Yoonju Jung
\IEEEcompsocitemizethanks{\IEEEcompsocthanksitem A. Heinecke is with the Department of Mathematics and Yale-NUS College, National University of Singapore, Singapore, 138527. 
E-mail: andreas.heinecke@yale-nus.edu.sg
\IEEEcompsocthanksitem E. Mayer is with the Division of Humanities, Yale-NUS College, Singapore, 138527. E-mail: emanuel.mayer@yale-nus.edu.sg
\IEEEcompsocthanksitem A. Natarajan is with the University of Cambridge. E-mail: abhinav.natarajan@u.yale-nus.edu.sg
\IEEEcompsocthanksitem Y. Jung is with Yale-NUS College, Singapore, 138527.}
\thanks{}}
\IEEEtitleabstractindextext{%
\begin{abstract}
Die analysis is an essential numismatic method, and an important tool of ancient economic history. Yet, manual die studies are too labor-intensive to comprehensively study large coinages such as those of the Roman Empire. We address this problem by proposing a model for unsupervised computational die analysis, which can reduce the time investment necessary for large-scale die studies by several orders of magnitude, in many cases from years to weeks. From a computer vision viewpoint, die studies present a challenging unsupervised clustering problem, because they involve an unknown and large number of highly similar semantic classes of imbalanced sizes. We address these issues through determining dissimilarities between coin faces derived from specifically devised Gaussian process-based keypoint features in a Bayesian distance clustering framework. The efficacy of our method is demonstrated through an analysis of 1\,135 Roman silver coins struck between 64-66~C.E.. 
\end{abstract}

}

\maketitle

\IEEEdisplaynontitleabstractindextext

%
\IEEEpeerreviewmaketitle

\ifCLASSOPTIONcompsoc
\IEEEraisesectionheading{\section{Introduction}\label{sec:introduction}}
\else
\section{Introduction}\label{sec:introduction}
\fi

%
%
%
%
\IEEEPARstart{A}{ncient} coins were minted from hand-engraved dies. With very few exceptions, the dies themselves are lost, but their impressions survive on the coins that were struck by them. When working with a sufficiently large sample, it is possible to determine through visual analysis how many individual dies were used to strike a chronologically discreet series of coins. This, in turn, allows for estimating the output of the mint that produced them – often on an annual basis, as in the case of dated Roman coins \cite{esty2011geometric}. In theory, this could provide ancient historians with much needed statistical data on an important component of the money supply and the fiscal regime of ancient states \cite{de1995calculating}. Yet in practice, die studies are too labor-intensive to systematically tackle the coinages of large polities such as the Roman Empire, which often commissioned hundreds of dies in a single year to mint millions of coins of exactly the same design.

Die studies are immensely time-consuming because the fine-grained class structure requires an enormous number of pairwise comparisons between the coins in each sample, first for the front faces (obverses) and then for the back faces (reverses). Ideally, every obverse is at least once compared to every other obverse, and every reverse to every other reverse. The number of required comparisons for a sample of $N$ images is thus at least $N(N-1)/2$. For instance, a study of 150 obverses would require more than ten thousand visual comparisons, whereas a study of 1\,500 obverses would already require more than 1.1 million comparisons. This quadratic growth in the number of comparisons makes manual die studies of substantial size unfeasible. Aargaard and M\"{a}rcher~\cite{aagaard2015microscope} consider analyzing 1\,000 coins a year’s worth of work, even with the help of a microscope drawing tube. Van Alfen~\cite{vanAlfen} estimates that a die study of the 60\,000 available 5th century  Athenian ``owl'' tetradrachms would consume a lifetime.

\begin{figure}[!t]
\centering
\subcaptionbox{Die 1\label{Fig:1a}}{\includegraphics[width=0.24\textwidth]{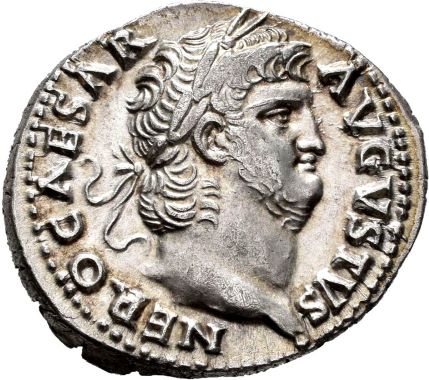}}
\subcaptionbox{Die 1\label{Fig:1b}}{\includegraphics[width=0.24\textwidth]{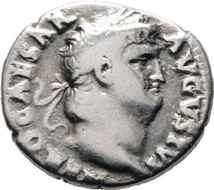}}
\subcaptionbox{Die 1\label{Fig:1c}}{\includegraphics[width=0.24\textwidth]{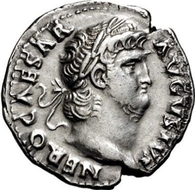}}
\subcaptionbox{Die 2\label{Fig:1d}}{\includegraphics[width=0.24\textwidth]{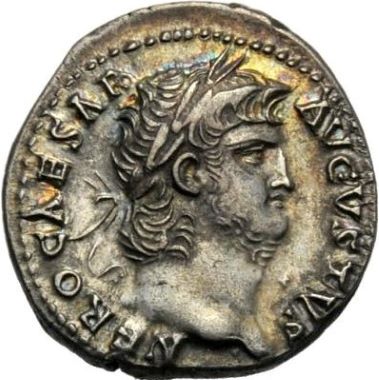}}
\caption{Variability and similarity in four Neronian denarii of the same obverse type struck from two different dies. Compared to (a), the coin shown in (b) is badly worn. The legend lettering in (c) is less clearly defined because of die deterioration; the horizontal artefact between the forehead and the letter A in Augustus - also seen in (b) - is the result of a die break. Die 2 is very similar to Die 1, but differs in the arrangement of individual locks and the position of the fillet vis-\`{a}-vis the lettering of the legend. Both dies were probably engraved by the same artisan.}
\label{Fig:DifferentOrSame}
\end{figure}

This enormous time investment is compounded by the difficulties posed by the material itself. Two coins struck from the same die often come in radically different states of preservation (Fig.~\ref{Fig:1a},\ref{Fig:1b}). Die deterioration during the striking process also affects the coin image (Fig.~\ref{Fig:1c}). Finally, dies made by the same engraver were often so similar that the coins struck by them are hard to tell apart (Fig.~\ref{Fig:1d}). These difficulties have discouraged many ancient historians and archaeologists from conducting die studies. As pointed out by de Callata\"{y}~\cite{de2011quantifying}, it constitutes a considerable career risk to embark on an immensely time-consuming numismatic project that may only result in a useful dataset but not yield a historically exciting result.

We address this dilemma by introducing a computational method to separate unlabelled target images of ancient coins by the dies from which they were struck. The dies correspond to unknown semantic classes, which have to be predicted along with their total number in an unsupervised manner. This is achieved through a clustering algorithm on the basis of intra-class similarities and cross-class differences between the images. The unsupervised workflow we propose has the potential to reduce the number of required visual comparisons in die studies by several orders of magnitude, and, in many cases, to collapse what used to be the work of months and years into the space of just a few days to weeks.

The goal of automating die studies has been advocated for some time \cite{howgego2005}. The American Numismatic Society has taken a pioneering role and developed a Computer Aided Die Study (CADS) software \cite{vanAlfen,taylor2020computer}. In contrast to the model presented here, CADS is intended to provide a visual pre-sorting tool for semi-manual hierarchical clustering of digital images. It does not explicitly partition a sample according to dies (see Sec.~\ref{subsec:CADS} for details on differences between our approach and CADS). The difficulty of computer assisted die studies is further underlined by Horache et al.~\cite{horache2021riedones3d}, who consider supervised deep representation clustering (based on~\cite{Choy2019FCGF}) for 3D-scans of a hoard find of Celtic coinage. The challenges of creating such a high-quality dataset lie not so much in the availability of a 3D-scanning device, as in the access to the coinage for scanning. Corpora of coins from non-hoard-finds that are of historical interest are, usually, not available in a single place. For instance, Tolksdorf et al.~\cite{Tolksdorf2017_37coins} can only assemble a dataset of 37 specimen of Roman imperial coinage from different archaeological sites for 3D-scanning, in order to address questions of movements of Roman legions. Moreover, while 3D-scans are able to improve the quality of the input data with respect to corrosion, and can factor out the problem of uneven lighting during image acquisition, details lost through wear deterioration, such as in Fig.~\ref{Fig:1a} versus~\ref{Fig:1b}, cannot be recovered (see Sec.~\ref{sec:comparison} for further details on quantitative and conceptual differences of \cite{horache2021riedones3d} to our approach).
To the best of our knowledge, besides our proposal there is currently no unsupervised method capable of end-to-end clustering of coins by dies. Indeed, an unsupervised method is confronted with complex images, exhibiting cross-class visual discrepancies and intra-class affinities that are much lower than in most natural image clustering tasks. In the absence of ground-truth labels, positive cluster pairs have to be identified purely based on visual transformations without information on visual discrepancy within classes or affinity across classes. The lack of guiding high-level knowledge about the global solution also implies that different, undesired clustering solutions, say by preservation state, can make sense of the input data. 

The sample coinage chosen for the demonstration of our method are denarii struck between 64-66~C.E., immediately after the Roman emperor Nero had reduced both the weight and, for denarii, the precious metal content of Rome’s gold and silver coinage. Nero’s so-called ``post-reform'' denarii are particularly salient for discussing issues of quantification. Since Mommsen~\cite{mommsen1860geschichte} theorized in 1860 that Nero’s debasement was motivated by fiscal difficulties and an attempt to harmonize the value of the denarius with eastern silver currencies, there has been a lively debate about the role of bullion coinage and of minting for meeting Rome’s fiscal needs. So far, this question has been addressed without meaningful estimates of how many coins Nero minted, which is, however, the key to historically sound answers. A comprehensive historical and numismatic analysis of this material will be presented in a die study of Nero's entire bullion coinage in the relevant specialist literature.

\begin{figure*}[!t]
\centering
\subcaptionbox{\emph{Extremely fine} ($>90\%$ design remaining)}{\includegraphics[width=0.33\textwidth]{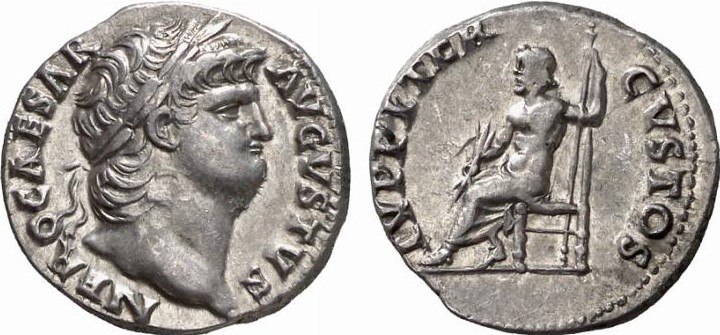}}
\subcaptionbox{\emph{Very fine} ($>75\%$ design remaining)}{\includegraphics[width=0.33\textwidth]{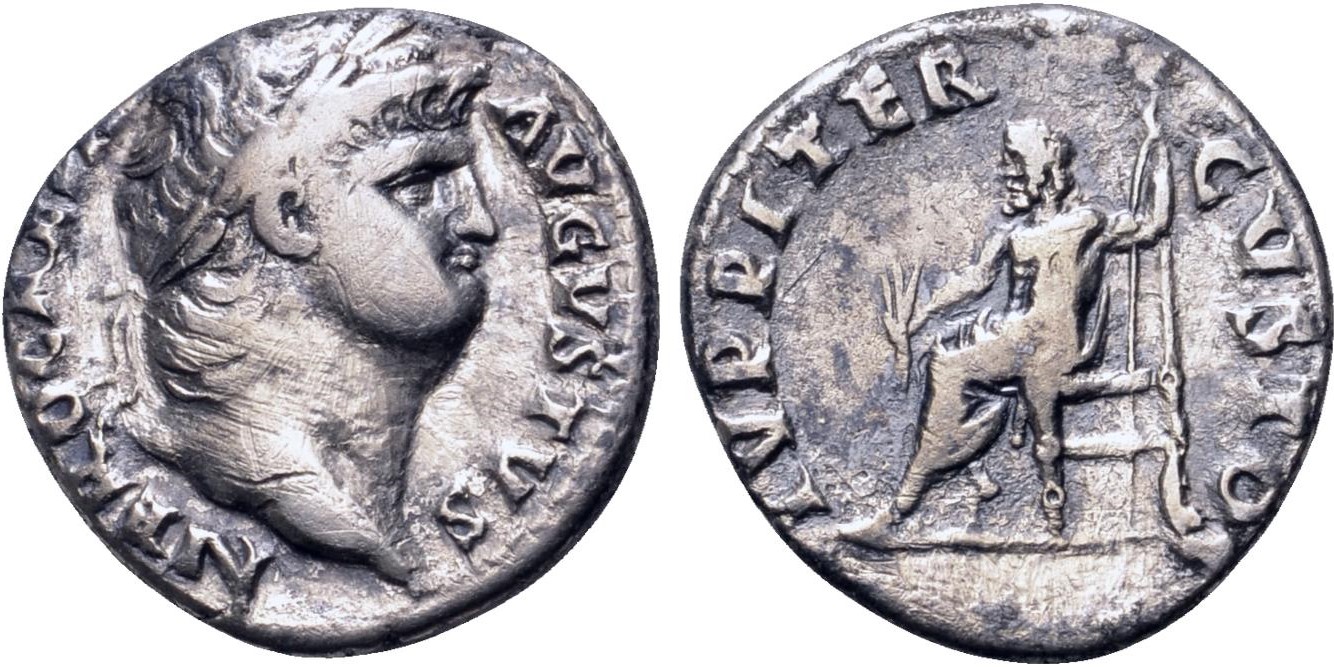}}
\subcaptionbox{\emph{Good} ($>10\%$ design remaining) (\copyright \,KHM)}{\includegraphics[width=0.33\textwidth]{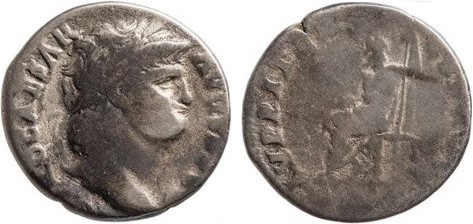}}
\caption{Preservation variability: Three denarii of differing preservation grades struck from the same obverse and three different reverse dies. 
}
\label{Fig:CoinPreservation}
\end{figure*}

The literature highlighted in Sec.~\ref{Sec:RelatedWork} below on computer vision methods in numismatics suggests that automated type classification of ancient coins is already a formidable task. Challenges only aggregate when attempting an automated unsupervised clustering method for die analysis: semantic classes (corresponding to dies) are unknown, fine-grained, and of imbalanced sizes (usually in the range of 1-30); labeled reference data are unavailable; derivation of discriminative features capable of promoting sufficient cluster compactness and separation is demanding, because of the intricate nature of the target images: both the variability for coins struck from the same die \emph{and} the similarity between coins struck from different dies are high. Finally, image acquisition characteristics and image qualities vary.  
The method proposed in this article is tailored to address these difficulties through an unsupervised workflow, comprising of
\begin{itemize}
    \item specifically devised keypoint feature extraction via a Gaussian process, and 
    \item Bayesian distance microclustering based on pairwise dissimilarities derived from the extracted features.
\end{itemize}
The Gaussian process based keypoint extraction we introduce for feature detection is customized for die analysis, which differs in several aspects from other computer vision tasks, such as object recognition or tracking, for which Difference-of-Gaussian based methods have been developed.
Our uncertainty based construction, and the subsequent matching of the keypoints between image pairs, is inspired by the work of \cite{gao2019gaussian1}, who consider Gaussian process landmarking on manifolds for analysis and comparison of three-dimensional anatomical shapes in evolutionary biology \cite{gao2019gaussian2}.
We utilize the pairwise dissimilarities calculated from the Gaussian process based features for Bayesian distance based microclustering \cite{natarajan2021repulsion}. In the Bayesian framework, the number of dies used to strike the sample, a key unknown \cite{lindhe2021variational}, is an object of inference, to be predicted in conjunction with the die labels. 
Its probabilistic foundation sets it apart from other distance-based clustering algorithms, such as hierarchical clustering employed by CADS, the graph-based clustering of~\cite{horache2021riedones3d}, or partitional approaches like $k$-means. The use of a microclustering prior addresses the fact that in ancient numismatics the frequency of surviving coins per die is generally  small, which results in relatively small class sizes even for large samples. A repulsion component in the prior further promotes separation of coins from similar dies (in a sense, \emph{close-by} clusters) that contribute to intra-class distances being of comparable sizes as adjacent cross-class distances.


\section{Related work} \label{Sec:RelatedWork}%
Previous to the above mentioned pre-sorting software CADS~\cite{taylor2020computer} and the supervised clustering in~\cite{horache2021riedones3d}, computer vision techniques in numismatics have been used for tasks such as ancient coin classification by numismatic type or issue \cite{kampel2008recognizing,LiuSIFTflow2011,Zambanini2013,Zambanini2013Illumination,Anwar2015,aslan2020}, coin identification \cite{Huber2012AutomaticCC}, and high-volume type classification of contemporary coins~\cite{Nolle2003}. While type classification is an interesting computer vision problem, it only is of limited value for historical research, because most available coin images already exist in the form of labelled data (Sec.~\ref{Sec:Data}).
Type classification methods cannot be easily adapted to computational die analysis. For instance, supervised learning approaches employed by these classification methods require reference images, or the knowledge of dies and their number in any given sample. 
This defeats the purpose of die analysis, the objective of which is the identification of yet unknown dies, i.e.\ semantic classes, as well as the determination of their number in the sample under study. In addition to this conceptual problem, most numismatic type classification methods mentioned above rely on off-the-shelf computer vision tools for the extraction and matching of local features, for example SIFT~\cite{lowe2004distinctive}, a variant of which is also used by CADS. Such Difference-of-Gaussian based methods were not designed for detecting subtle differences between highly similar objects, which is the main challenge in die analysis. Indeed, in the literature on type classification, they are used to distinguish between different types of numismatic portraits \cite{Arandjelovic2010}, different coin legends \cite{Kavelar2012}, different motifs \cite{Anwar2015motif}, or a combination of these \cite{Arandjelovic2012}. 
More recent approaches to type classification of ancient coins utilize supervised learning via convolutional neural networks \cite{Kim24,Kim2016,Schlag2017,Anwar2021Deep}, often after discarding badly preserved coins, which are common in ancient numismatics.  

\section{Data collection and description}\label{Sec:Data}

\begin{figure}[htb]
\centering
 \begin{subfigure}[b]{0.19\columnwidth}
        \centering
        \includegraphics[width=\columnwidth]{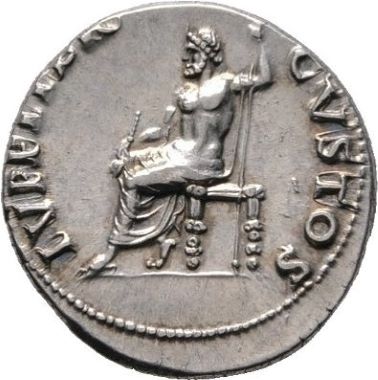}
        
        \includegraphics[width=\columnwidth]{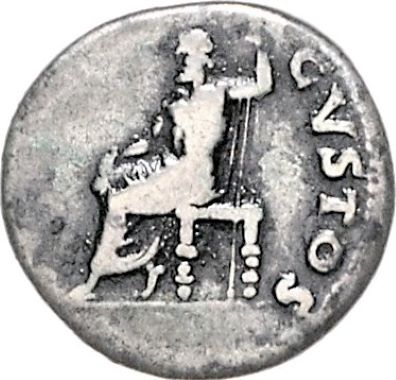}
        \caption*{\emph{RIC} I\textsuperscript{2} 53}
 \end{subfigure}
  \begin{subfigure}[b]{0.19\columnwidth}
        \centering
        \includegraphics[width=\columnwidth]{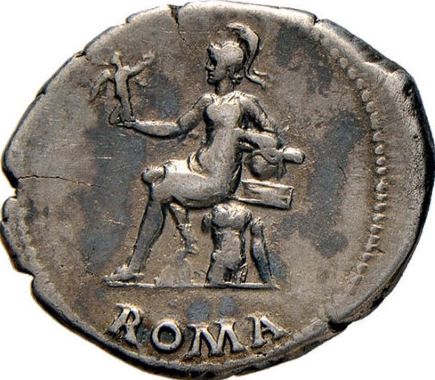}
        
        \includegraphics[width=\columnwidth]{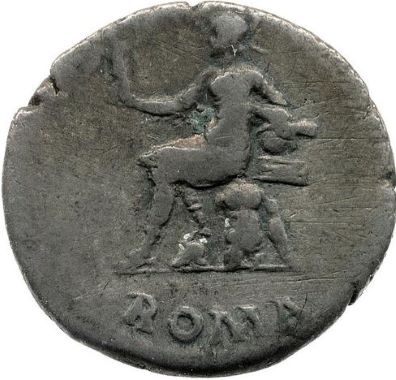}
        \caption*{\emph{RIC} I\textsuperscript{2} 55}
 \end{subfigure}
  \begin{subfigure}[b]{0.19\columnwidth}
        \centering
        \includegraphics[width=\columnwidth]{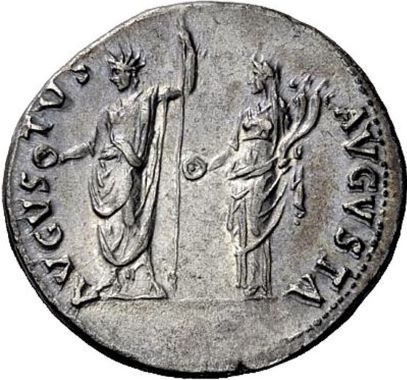}
        
        \includegraphics[width=\columnwidth]{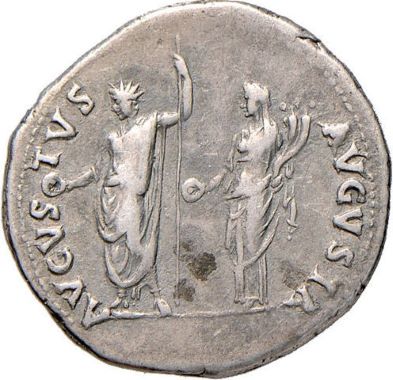}
        \caption*{\emph{RIC} I\textsuperscript{2} 57}
     \end{subfigure}
\begin{subfigure}[b]{0.19\columnwidth}
        \centering
        \includegraphics[width=\columnwidth]{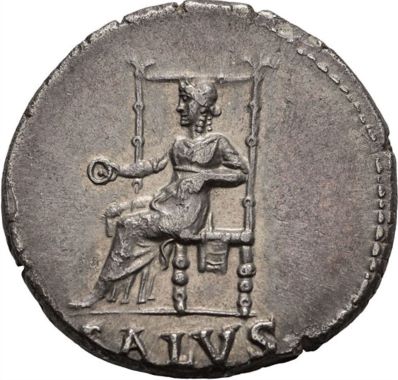}
        
        \includegraphics[width=\columnwidth]{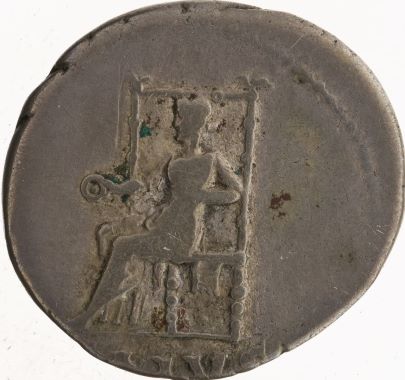}
        \caption*{\emph{RIC} I\textsuperscript{2} 60}
 \end{subfigure}
  \begin{subfigure}[b]{0.19\columnwidth}
        \centering
        \includegraphics[width=\columnwidth]{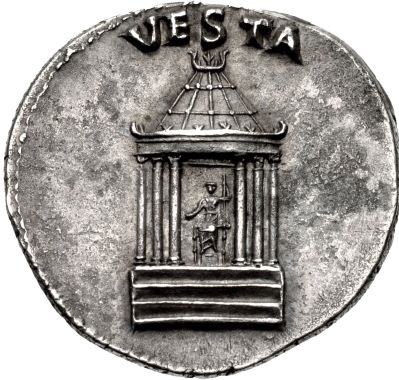}
        
        \includegraphics[width=\columnwidth]{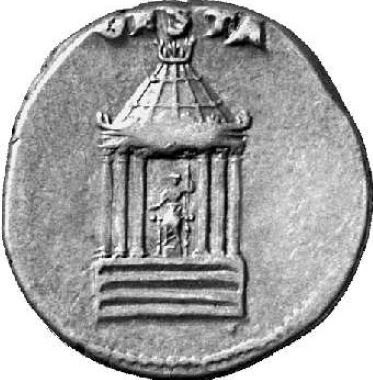}
        \caption*{\emph{RIC} I\textsuperscript{2} 62}
 \end{subfigure}
\caption{Illustration of the five reverse types under study. Displayed is, for each type, a well preserved and a poorly preserved specimen struck from the same die.}
\label{Fig:ReverseMotifs} 
\end{figure}

\begin{figure}[!ht]
\caption{Variability in images of the same coin: Auction lots advertising the same specimen in different years, with photographs taken with different lighting and published in different resolution. In addition, the displayed obverse acquired cabinet toning between 2003 and 2018, which is not visible in the 2019 photograph.}
\label{Fig:Identicals}
\centering
\subcaptionbox{2003\label{Fig:2a}}{\includegraphics[width=0.32\columnwidth]{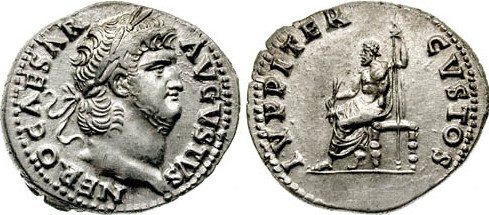}}
\subcaptionbox{2018\label{Fig:2b}}{\includegraphics[width=0.32\columnwidth]{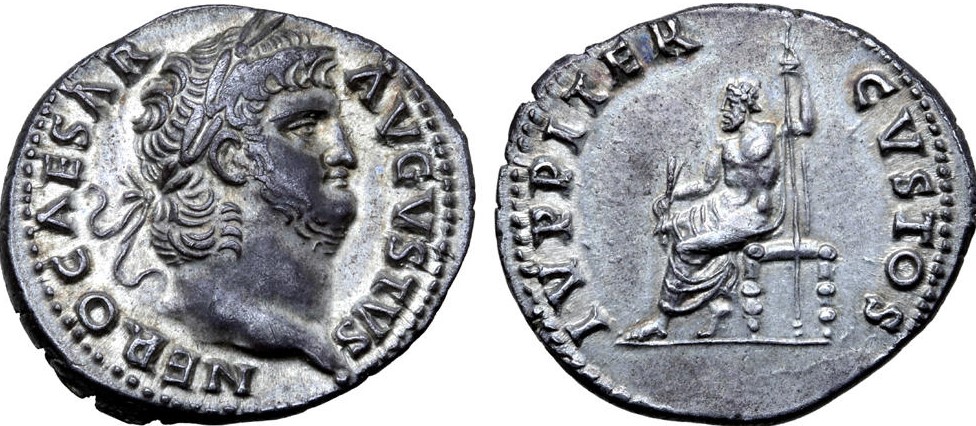}}
\subcaptionbox{2019\label{Fig:2c}}{\includegraphics[width=0.32\columnwidth]{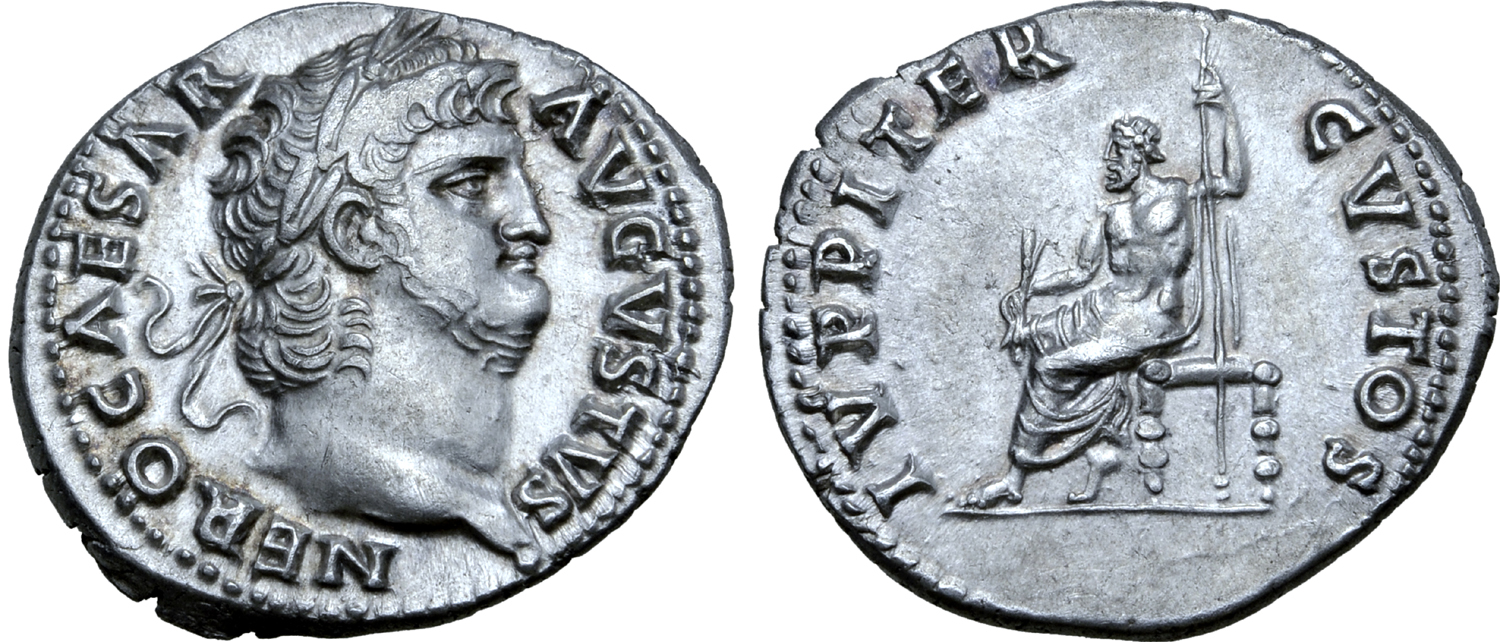}}
\end{figure}

Millions of coin images classified by numismatic type are available online, most importantly through CoinArchivesPro~\cite{CoinArchivesPro}, a sub\-scrip\-tion-\-based website catering to coin collectors, coin dealers, and numismatists that archives all individual coins from most major coin auction catalogues since 1996. The coin holdings of many public collections are accessible through online databases such as Online Coins of the Roman Empire (OCRE~\cite{ocre}). Numismatic paper publications, on the other hand, are particularly important for studying significant hoard finds, but are only available through specialist research libraries, and often only come with halftone photographs. Online databases archive coins according to the classification systems of academic catalogues such as Roman Imperial Coinage (RIC~\cite{sutherland2018roman}).
This allows for building a research sample quickly,
even though the results of any search must be checked for occasional mis-classifications. 
Auction catalogues contain by far the largest collection of ancient coinage, but a significant number of individual coins were offered up for auction more than once, and re-published with new photographs (Fig.~\ref{Fig:Identicals}). Consequently, any extensive collection effort will yield duplicate images of the same coin. Further, any such sample will vary considerably in image quality and illumination. Neither the auction catalogues nor the public collections contributing to sites such as OCRE photograph coins to a uniform standard.
The most significant challenge is, however, the variation in coin preservation (Fig.~\ref{Fig:CoinPreservation}). Few ancient coins have survived in excellent condition. When working with a large sample, the preservation of many specimens is poor. This also applies to the dataset considered in this article. We rely on the coin grading scales employed by the auction catalogues for preservation statistics (with missing grades provided by us) to gauge the performance of our model vis-à-vis the quality of the input data (see Sec.~\ref{Sec:Results}).
For a critical evaluation of this grading scale for research, we refer to~\cite{AZ2020}.

\begin{figure}[!ht]
\caption{Die combinations transcending \emph{RIC} classifications: Three denarii struck from one obverse die and three reverse dies with different designs. The mint of Rome routinely combined a single obverse die with reverse dies of different numismatic types. As a result, obverse dies would be over-counted if studied by the reverse-type based \emph{RIC} numbers.}
\label{Fig:Reverses}
\centering
\subcaptionbox{\emph{RIC} I\textsuperscript{2} 53}{\includegraphics[width=0.32\columnwidth]{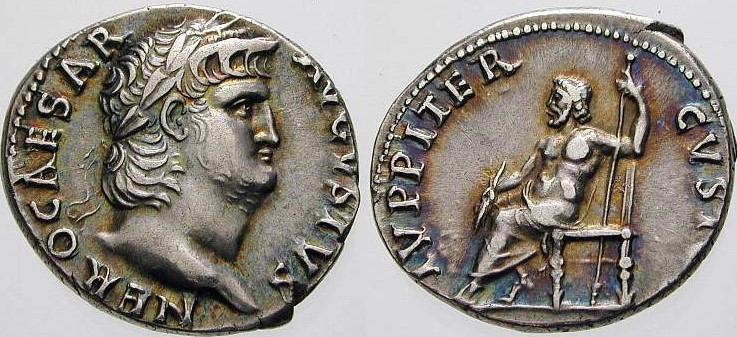}}
\subcaptionbox{\emph{RIC} I\textsuperscript{2} 60}{\includegraphics[width=0.32\columnwidth]{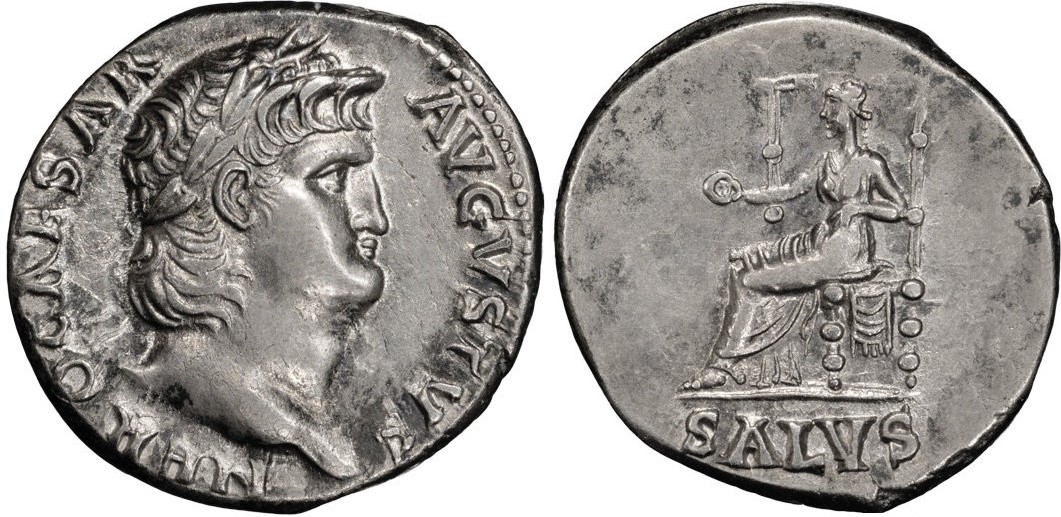}}
\subcaptionbox{\emph{RIC} I\textsuperscript{2} 62}{\includegraphics[width=0.32\columnwidth]{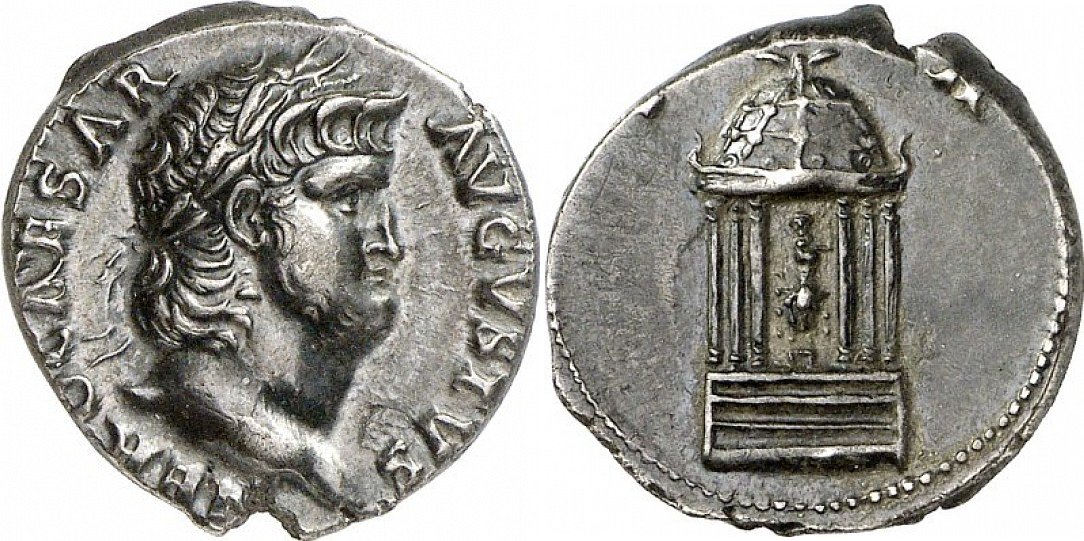}}
\end{figure}

For this article, 2\,866 images (1\,434 obverses, 1\,432 reverses) 
of ``post-reform'' Neronian denarii (\emph{RIC} I\textsuperscript{2}
53; 55; 45/57; 60; 62) were considered (see Fig.~\ref{Fig:ReverseMotifs} for an illustration of the reverse motifs). These images only represent 1\,135 coins. Nevertheless, all images were supplied to our computational method, as detecting duplicate images is a genuine problem in die analysis. While reverse types were studied separately, all obverses were studied as a single set, because the mint of Rome routinely combined obverse dies with reverse dies of several types (Fig.~\ref{Fig:Reverses}).

\section{Methodology}\label{Sec:Methods}

\begin{figure}[!t]
\centering
\includegraphics[width=0.24\textwidth]{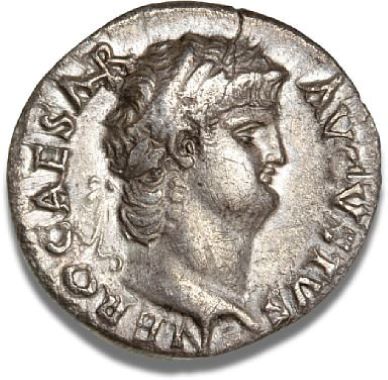}
\includegraphics[width=0.24\textwidth]{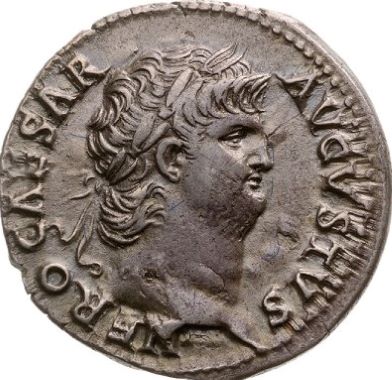}
\includegraphics[width=0.24\textwidth]{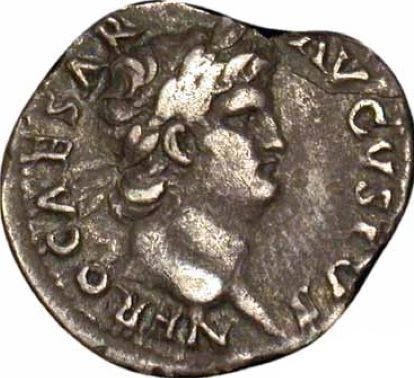}
\includegraphics[width=0.24\textwidth]{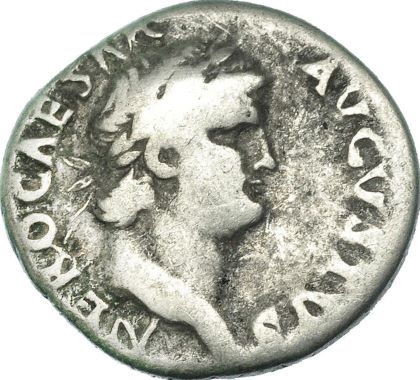}
\caption{Cross-class similarity: Obverses struck from four different dies, apparently engraved by two different artisans. The coin pairs in the top and bottom row share many minute and characteristic details such as the arrangement of the fillet of the laurel wreath. However, subtle differences in the arrangement of individual locks and letters identify them as struck from different dies. Poor preservation adds additional difficulty to die identification.}
\label{Fig:SimilarDies}
\end{figure}

In this section we present the details of our proposed method to automatically cluster ancient coin images according to the dies from which they were struck. 
The basis for cluster label assignments -- representing the die predictions -- is the definition of a dissimilarity measure for image pairs. These dissimilarities may be thought of as distances between images. Despite the intricate and challenging class structure, the dissimilarity measure has to enable discrimination by assigning a larger distance between coin faces struck from different dies than from the same die. The goal of the subsequent clustering step is to assign the same labels to image pairs with low dissimilarity. However, the nature of the input data for die studies in ancient numismatics presents several challenges that make off-the-shelf computer vision tools and clustering techniques not ideally suited for dissimilarity and label assignments, and require customized solutions.

\begin{figure}[!t]
\centering
 \begin{subfigure}[b]{0.24\textwidth}
\includegraphics[width=\textwidth]{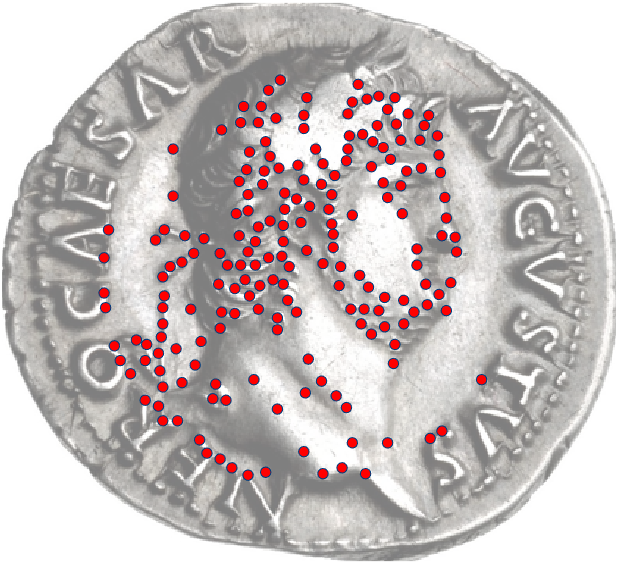}
\caption{GP-keypoints}
        \label{Fig:190GP}
 \end{subfigure}
 \begin{subfigure}[b]{0.24\textwidth}
\includegraphics[width=\textwidth]{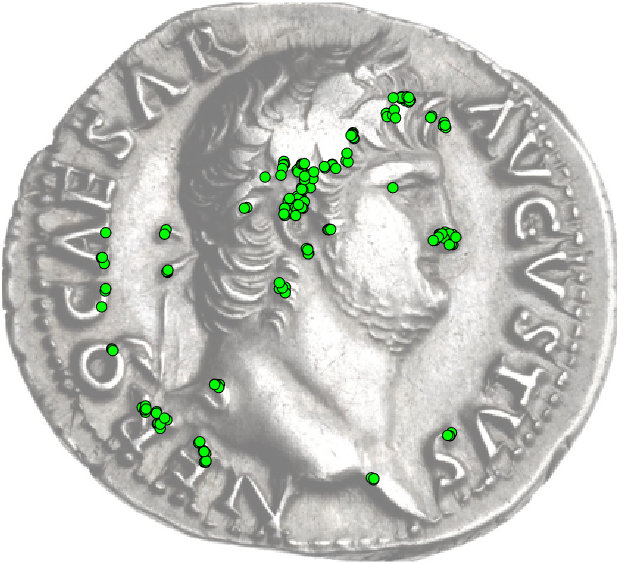}
 \caption{ORB-keypoints}
 \label{Fig:190ORB}
 \end{subfigure}
\caption{Visual comparison of 190 GP-keypoints  with the 190 strongest ORB-keypoints~\cite{rublee2011orb} for the same region of interest.
ORB-keypoints accumulate where individual keypoints represent features at different scales according to the underlying Difference-of-Gaussians scheme.
}
\label{Fig:GPvsORB}
\end{figure}

\begin{figure*}[!t]
\centering
    \begin{subfigure}[b]{0.495\textwidth}
        \centering
        \includegraphics[width=0.49\textwidth]{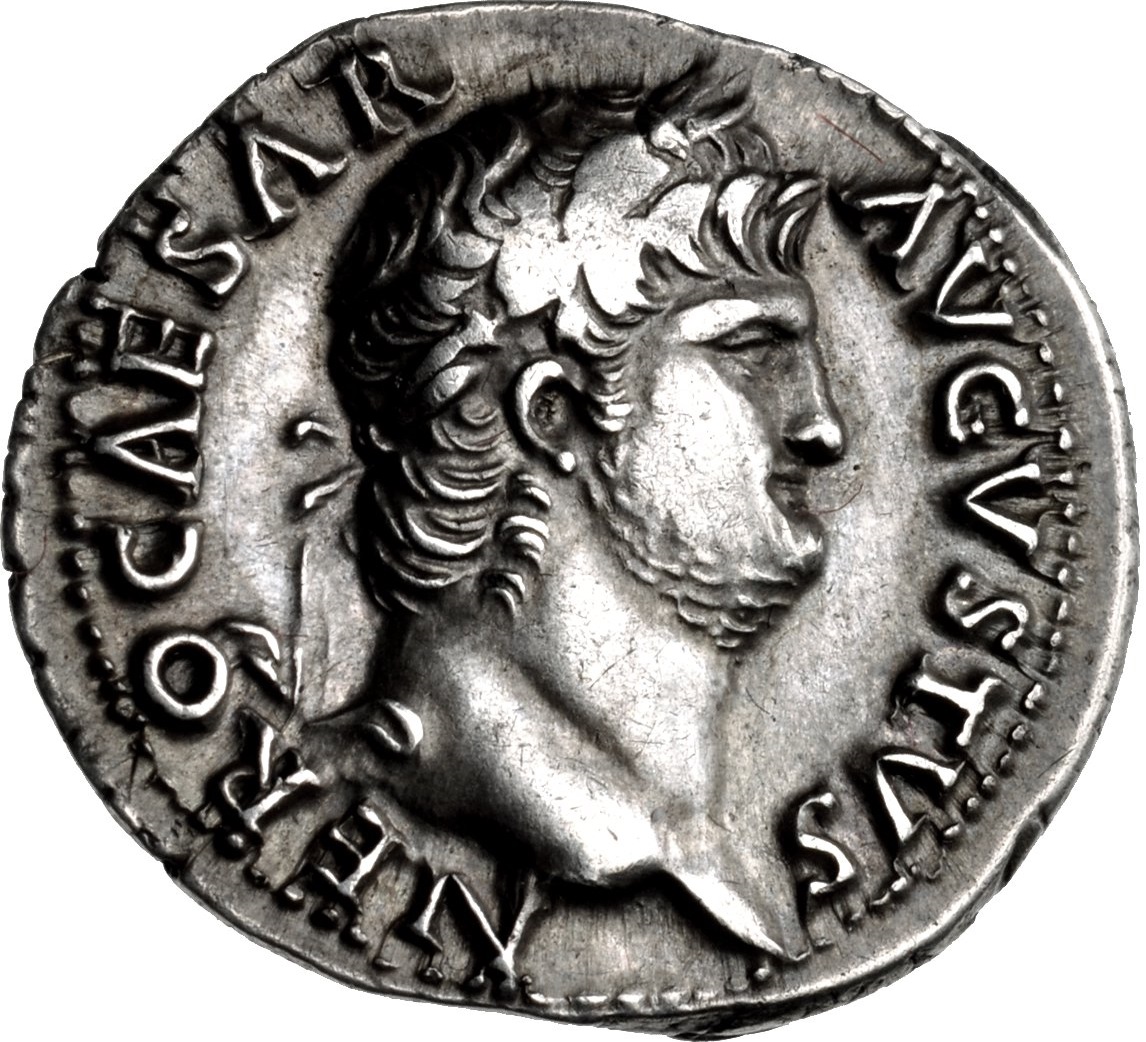}
        \includegraphics[width=0.49\textwidth]{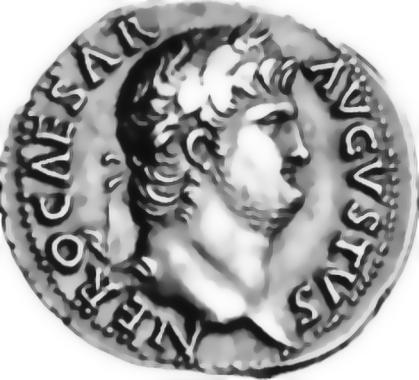}
        \caption{Raw and pre-processed obverse.}
        \label{Fig:preprocessing}
    \end{subfigure}
    \begin{subfigure}[b]{0.495\textwidth}
        \centering
        \includegraphics[width=0.49\textwidth]{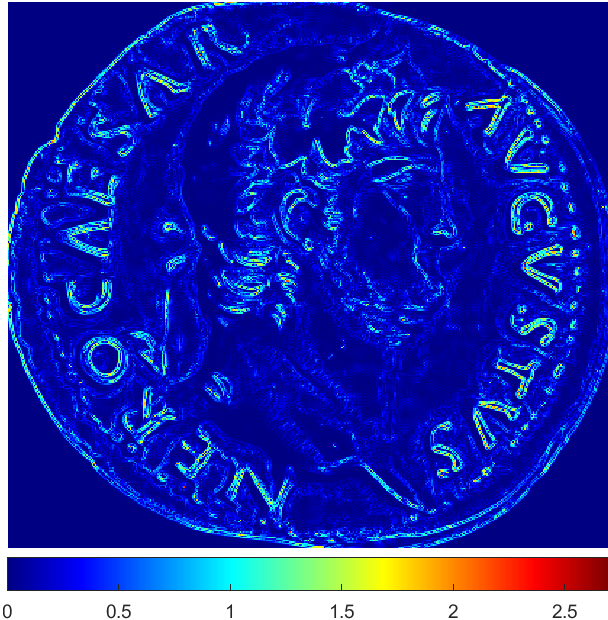}
        \includegraphics[width=0.49\textwidth]{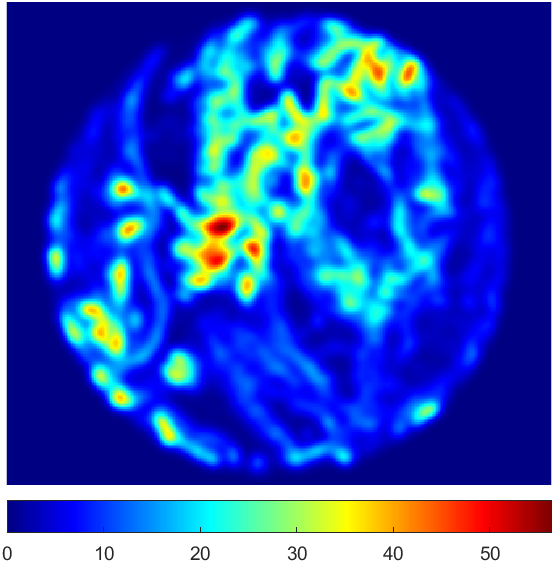}
        \caption{Laplace filtering and re-weighted covariance matrix.}
        \label{Fig:LMprior}
    \end{subfigure}
    
     \begin{subfigure}[b]{0.33\textwidth}
        \centering   
\includegraphics[width=0.49\textwidth]{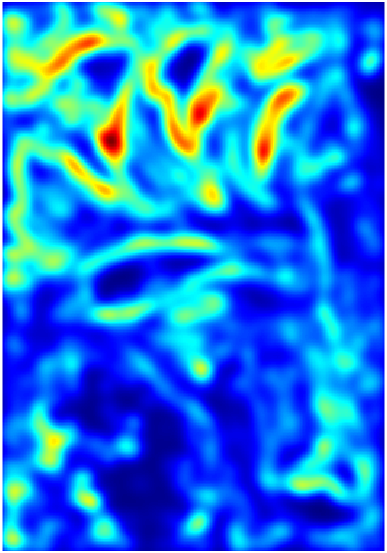}
\includegraphics[width=0.49\textwidth]{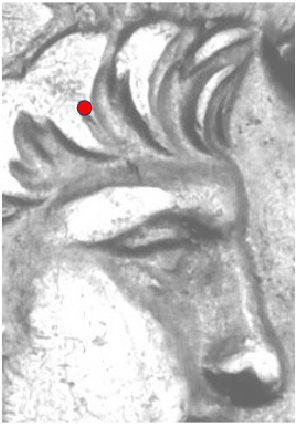}
        \caption{$\bSigma_1$ and choice of keypoint $\bxi_1$.}
        \label{Fig:5c}
    \end{subfigure}
    \begin{subfigure}[b]{0.33\textwidth}
        \centering
\includegraphics[width=0.49\textwidth]{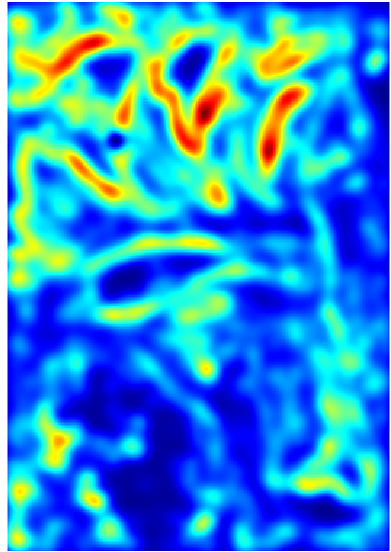}
\includegraphics[width=0.49\textwidth]{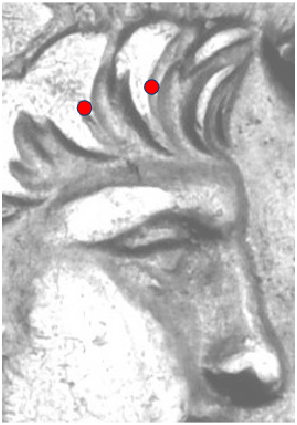}
        \caption{$\widetilde{\bSigma}_2$ and choice of keypoint $\bxi_2$.}
        \label{Fig:5d}
    \end{subfigure}
      \begin{subfigure}[b]{0.33\textwidth}
        \centering  
\includegraphics[width=0.49\textwidth]{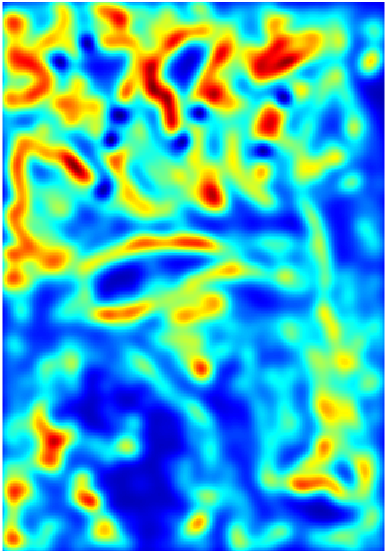}
\includegraphics[width=0.49\textwidth]{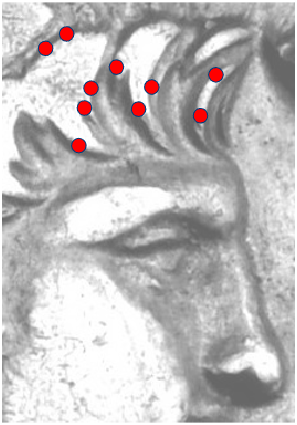}
        \caption{$\widetilde{\bSigma}_{10}$ and choice of keypoint $\bxi_{10}$.}
        \label{Fig:5e}
    \end{subfigure}
%
\caption{
Illustration of GP-keypoint construction: In (b), geometric information is extracted via isotropic Laplace filtering from the pre-processed obverse in (a). The re-weighted prior covariances for a circular region of interest are derived via convolution with a squared exponential kernel. Iterative choices of the first, second and tenth GP-keypoints as the pixel locations of maximal posterior uncertainty for the re-weighted Gaussian process, visualized as heatmaps for a selected region, are shown in (c-e).
}
\label{ImageToLMPrior}
\end{figure*}

To begin, determining distances that distinguish same-die from different-die image pairs requires a compact data representation of the images under study. In mathematical spaces of dimensions as large as the number of pixels in a digital coin image, there is little discrimination in Euclidean distances  -- all images are far apart. This general problem of featurization is analogous to the practice of visual die analysis. As in many computer vision methods, instead of pixel-wise comparisons, we compare images on the basis of keypoints, delineating salient features such as letters, locks, the shape of an eye, or the profile of a nose.
A significant challenge for predicting classes correctly is posed by the deterioration of dies during the minting process and the vastly differing preservation grades of the coins under study, in conjunction with the relative similarity between dies. In particular, badly preserved coin faces struck from different dies by the same engraver are difficult to label correctly (Fig.~\ref{Fig:SimilarDies}). 
This problem is addressed through a specifically devised feature extraction algorithm. The distances that we define on the the basis of custom-built keypoints (Fig.~\ref{Fig:190GP}, \eqref{ImageToLMPrior}) -- more precisely, of their relative spatial positions and correspondences -- overall succeed in distinguishing same-die and different-die pairs (Fig.~\ref{Fig:DistanceHistograms}). The challenge of small distance values between images belonging to similar -- but distinct --  dies/classes is addressed through the inclusion of both a cohesion and a repulsion component in the likelihood of the subsequent Bayesian clustering.

In addition to the fine-grained class structure, with subtle differences between classes, a problem is posed by the fact that the number of dies used to strike the sample is unknown. The Bayesian paradigm addresses this by considering the number of clusters as a parameter to be inferred from the dissimilarities. Finally, most numismatic samples from antiquity require microclustering. 
The microclustering assumption states that cluster sizes grow only sublinear with the total number of data points.
This property is crucial for die studies of numismatic samples from antiquity. Due to the low frequencies of surviving coins per die, cluster sizes are exceedingly small by the standards of most clustering or classification applications -- with rare exceptions of particularly large hoard finds mostly containing newly minted coins of the same numismatic type. Finally, class sizes are rather imbalanced, with many singleton classes and ranging to several tens of images.

In summary, the challenges posed by the numismatic data are addressed in the following automated workflow:
\begin{itemize}
    \item[(o)] Pre-processing,
    \item[(i)] Gaussian process keypoint extraction,
    \item[(ii)] Pairwise dissimilarity calculation,
    \item[(iii)] Bayesian distance microclustering.
\end{itemize}

The main purpose of the pre-processing step is to reduce noise, which is either introduced during image acquisition, for instance due to low illumination, or stemming from irregularities in the field, i.e., the background area of a coin image without design or lettering. Noise-reduction is achieved through a slight ``cartoonification'' of the images (Fig.~\ref{Fig:preprocessing}).
After converting all images to grey-scale and resizing to uniform height, total-variation regularization \cite{rudin1992nonlinear} is applied to each image, followed by contrast limited adaptive histogram equalization \cite{pizer1987adaptive} to enhance contrast and definition. Potential artefacts introduced during this enhancement are reduced through a second total-variation regularization.

In step (i), outlined in further detail in Sec.~\ref{subsec:GP}, keypoints are extracted for each coin face via a Gaussian process (GP). The standard regression application of a GP is to learn an unknown function from its values at given sample locations, at the same time providing uncertainties for function-value predictions away from the sample locations. By contrast, we use the covariance of a GP to iteratively choose for a known function -- the pre-processed coin image -- sample locations that, at every step of the iteration, have maximal uncertainty when modelling the function from the previously chosen sample locations via the GP -- i.e., for which querying the corresponding function value results in the largest possible information gain \cite{liang2015landmarking,Williams2006gaussian,cohn1996active,kapoor2007active}. To incorporate relevant information about the function beyond its values at previously chosen locations into the covariance structure describing the uncertainties, we use edge-relief information of the coin image to derive a data-driven covariance kernel. This ``re-weighting'' of a standard kernel with information of the function of interest is inspired by Gao et al.~\cite{gao2019gaussian2,gao2019gaussian1}, who use this method of uncertainty sampling of GP active learning to chose landmarks on three-dimensional surfaces of fossil molars from curvature information. From a different point of view, our method can also be described as iteratively choosing keypoint locations based on a matrix containing relief edge information of the coin image (Fig.~\ref{Fig:LMprior}): Its convolution with a squared exponential kernel function results in a ``blurring'' that combines at each pixel position information from its neighborhood, and corresponds to the prior covariances of the re-weighted GP uncertainties before choosing any sample locations. At each step of the iteration we then choose the location of maximal uncertainty, after which the uncertainty is locally shrunk towards zero in accordance with the information gained from the new sample. The updated matrix represents the posterior covariance structure of the re-weighted GP after choosing an additional keypoint, and the procedure is repeated (Fig.~\ref{Fig:5c}--\ref{Fig:5e}).

In steps (ii) and (iii), further detailed in Sec.~\ref{subsec:DistAndCluster}, distances between images are calculated on the basis of corresponding GP-keypoints that are matched using local gradient information and relative global positions. Dissimilarities between images are defined using the number of matching keypoints and the overall correspondence of their relative positions. We utilized parallel computing resources to calculate all pairwise dissimilarities, since the number of pairs is quadratic in the number of images. Clustering is then based on the dissimilarity scores via a chaperones algorithm implementation of the Bayesian distance microclustering model of \cite{natarajan2021repulsion}. 
Expert domain knowledge can be used to inform prior hyper-parameters. It is possible to include an expected mean and variance for the die frequencies, providing some amount of prior information on the total number of classes and range of class sizes. This can be helpful when it is clear from the outset that the number of singleton dies will be very high (e.g., due to a small sample size), or that the sample is suspected to contain a large number of coins from the same die (e.g., for certain hoard-finds). 


\subsection{Gaussian process keypoints}\label{subsec:GP}

In a visual die analysis, key features of individual coin faces are often identified when comparing pairs of coins. By contrast, to derive GP-keypoints, each pre-processed gray-scale image is considered individually as a scalar-valued function on its two-dimensional discrete pixel domain $\D$. Geometric information are first extracted by considering the pointwise absolute value of an isotropic filtering with a Laplace kernel. Alternatively, such information may be extracted using a Sobel filter, or high-pass wavelet coefficients. 
During the striking process, dies deteriorated faster near the periphery than in the centre of the flan, affecting the lettering of the coin legend first. Further, badly worn or mis-struck coins often retain almost no legend. However, the position of letters relative to the portrait or motif can provide important information to distinguish dies. The particular shapes of the coin circumference edges, on the other hand, bear no information about the dies used for striking. We therefore only retain the extracted geometric information inside a circular region that excludes the coin circumference edges, but contains much of the legend information (Fig.~\ref{Fig:LMprior}). Adapting the idea of \cite{gao2019gaussian1}, we use the resulting weights $\omega\in\R^{\D}$ to re-weight a discrete squared exponential kernel with lengthscale $\ell$, given by
\begin{align*}
    k_{\ell}(\bx,\by) := \exp\left(-\tfrac{\|\bx-\by\|^2}{2\ell^2}\right)
    \text{ for } \bx,\by\in \D,
\end{align*}
with geometric information, and use it to define the covariance structure of a GP. The particular form of re-weighting can be motivated by the reproducing property 
\begin{align}\label{Eq:ContReprProp}
  k_{\ell}(\bx,\by) \propto \int_{\R^2}  k_{\ell/2}(\bx,\bz) k_{\ell/2}(\bz,\by) \,\mathrm{d}\bz
  \text{ for } \bx,\by\in \R^2
\end{align}
that the squared exponential kernel possesses when considered on continuous two-dimensional domain. The re-weighted discrete kernel is defined by introducing the weights into the discrete analogue of \eqref{Eq:ContReprProp}
(see~\cite{gao2019gaussian1}, Eq.(3.8)), letting
\begin{align*}
    k_{\ell}^{\omega}(\bx,\by) := \sum_{\bz\in \D} k_{\ell/2}(\bx,\bz)\omega(\bz)k_{\ell/2}(\bz,\by)
    \text{ for } \bx,\by\in \D.
\end{align*}
The non-negativity of the weights guarantees the positive definiteness of the re-weighted kernel $k_{\ell}^{\omega}$, which is used to define the covariance structure of a GP. The prior covariances $(k_{\ell}^{\omega}(\bx,\bx))_{\bx\in\D}$ of this GP are given by the discrete convolution
\begin{align*}
\bSigma_1:=\omega * k_{\ell}(\cdot,(0,0))\in \R^{\D} 
\end{align*}
of the relief edge weights with the discrete squared exponential kernel (Fig.~\ref{Fig:LMprior}).
Given function values at $n$ points $\bxi_1,\ldots,\bxi_n$, the conditional covariance matrix $\bSigma_{n+1}\in\R^{\D}$ is a Schur-complement with entry indexed by $\bx\in \D$ given by (e.g.,\cite{Williams2006gaussian})
\begin{multline}\label{Eq:PosteriorExact}
    \bSigma_{n+1}(\bx) :=  k_{\ell}^{\omega}(\bx,\bx)  -  \\
   \left(\begin{smallmatrix}
      k_{\ell}^{\omega}(\bx,\bxi_1)\\
      \vdots\\
      k_{\ell}^{\omega}(\bx,\bxi_n)
    \end{smallmatrix}\right)^\top
    \left(\begin{smallmatrix}
      k_{\ell}^{\omega}(\bxi_1,\bxi_1)& \cdots& k_{\ell}^{\omega}(\bxi_1,\bxi_n)\\
      \vdots & & \vdots\\
    k_{\ell}^{\omega}(\bxi_n,\bxi_1)& \cdots& k_{\ell}^{\omega}(\bxi_n,\bxi_n)\\
    \end{smallmatrix}\right)^{-1}
    \left(\begin{smallmatrix}
      k_{\ell}^{\omega}(\bx,\bxi_1)\\
      \vdots\\
      k_{\ell}^{\omega}(\bx,\bxi_n)
    \end{smallmatrix}\right).
\end{multline}
GP-keypoints $\bxi_1,\bxi_2,\ldots,$ are iteratively chosen as
\begin{align*}
    \bxi_{n}:=\argmax_{\bx\in \D}\bSigma_{n}(\bx),
\end{align*}
i.e., at each step the selected keypoint is the most informative location, based on the uncertainty modelled by the re-weighted GP. For large images, the calculation of \eqref{Eq:PosteriorExact} is computationally expensive.
However, due to the rapid decay of the squared exponential kernel, $k_{\ell}^{\omega}(\bx,\by)$ is essentially zero unless the distance between the pixel-positions $\bx$ and $\by$ is small. Given the small lengthscale $\ell$ we used in our implementation, we therefore approximated \eqref{Eq:PosteriorExact} for computational efficiency by setting $k_{\ell}^{\omega}(\bxi_i,\bxi_j)$ to zero whenever $i\neq j$, resulting in the posterior covariance approximations 
\begin{align}\label{Eq:PosteriorApprox}
    \widetilde{\bSigma}_{n+1}:=\widetilde{\bSigma}_n-\left(\bSigma_1(\bxi_n)\right)^{-1} \left(k_{\ell}^{\omega}(\cdot,\bxi_n)\right)^2
    \text{ for } n\geq 1,
\end{align}
with $\widetilde{\bSigma}_1:=\bSigma_1$ (Fig.~\ref{Fig:5c}--\ref{Fig:5e}). 
Our numerical experiments indicated no notable difference between the keypoints derived from \eqref{Eq:PosteriorExact} and its approximation \eqref{Eq:PosteriorApprox}.

\subsection{Dissimilarities and microclustering}\label{subsec:DistAndCluster}%
\begin{figure}[!t]
\centering
\includegraphics[width=\columnwidth]{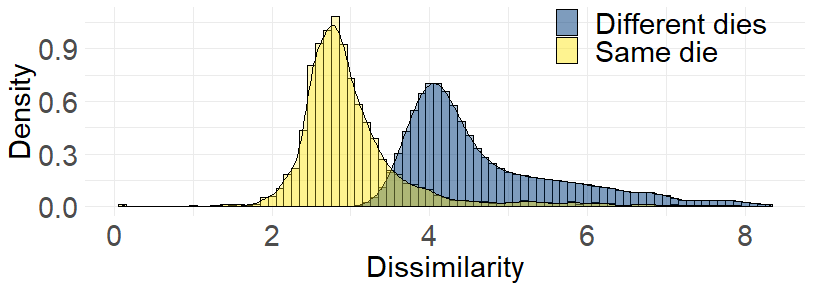}
\caption{Distributions of dissimilarity scores for all obverse pairs, separated into same-die and different-die pairs. 
The mean dissimilarity of the same-die pairs is clearly below that of the different-die pairs. Yet, same-die pairs with starkly differing preservation grades may still have a larger dissimilarity score than different-die pairs of well preserved coins, or coins struck from different dies by the same engraver.
(Relative frequencies are shown for visualization, since different-die pairs outnumber same-die pairs by three orders of magnitude.)}
\label{Fig:DistanceHistograms}
\end{figure}

Based on the GP-keypoints, a dissimilarity score for each image pair $(i,j)$ (for $i=1,\ldots,N$ and $j=i+1,\ldots,N$) is calculated as follows.
In a pre-matching step, each keypoint of image $i$ is associated with its best matching keypoint in image $j$ according to  neighborhood gradient information. Keypoints without suitable pre-matches are discarded. We used VLFeat \cite{vedaldi08vlfeat}, which, for custom keypoints, allows calculating both descriptor vectors summarizing local gradient information and subsequent matches.
To further increase the geometric and semantic consistency of the subset of matched GP-keypoints, and to remove outliers, the number of pairs is further reduced via low-distortion correspondence filtering \cite{lipman2014feature,gao2019gaussian2,Lofberg2004}, retaining the largest subset of matched keypoints, such that the keypoints of image $i$ and $j$ can be aligned via a non-rigid deformation with global distortion below a prescribed bound.
For the $n_{ij}$ matched keypoints of image pair $(i,j)$ that remain after this process, we calculate the Procrustes distance $p_{ij}$ \cite{gower1975generalized,gower2004procrustes}, i.e., the minimal total Euclidean distance between the matched keypoints of image $i$ and $j$ over all possible rigid motions. When $n_{ij} \leq 2$ (indicating a large dissimilarity) the Procrustes distance is always zero, so we set $p_{ij}$ to a large value in this case (informed by the maximum Procrustes distance across the dataset). If, as in our sample, the global rotation of the coins varies only slightly across all images, a threshold on the rotation angle of the Procrustes transformation can be used to further discard semantically meaningless matches. After shifting and rescaling the dissimilarity measures $1/n_{ij}$ and $\log p_{ij}$ over all pairs $(i,j)$ to the same range, we use their sum $d_{ij}$ as distance between the images $i$ and $j$. Fig.~\ref{Fig:DistanceHistograms} shows the distribution of these distances for all image pairs in our obverse dataset, separated by same-die and different-die pairs. The mean of the same-die distances is clearly smaller than the mean of the different-die distances, though, besides outliers from the same-die pairs, there is a challenging region of overlap, e.g., due to distances between coin faces struck from highly similar dies being comparable to same-die distances.

The distances $d_{ij}$ are used to cluster the images via the Bayesian distance microclustering algorithm developed in \cite{natarajan2021repulsion}, to which we refer for further details.
The Bayesian distance based approach defines a likelihood directly on the  distances between observations \cite{DuanDunson2018}, representing a middle ground between model based clustering~\cite{Quintana2006}, which for high-dimensional data is computationally prohibitive, and distance-based clustering, such as hierarchical or $k$-means clustering which lack a probabilistic background. Indeed, an advantage of the Bayesian approach over the latter distance based methods is its joint estimation of the number of clusters from the data, and \cite{natarajan2021repulsion} shows that their method can outperform hierarchical and $k$-means clustering even when those are provided with the number of clusters (usually unknown in applications).

Given the large number of images to be clustered and the relatively small cluster sizes in numismatic data sets from antiquity (i.e., large number of clusters), the Gibbs sampling of the Markov Chain Monte Carlo (MCMC) algorithm outlined in \cite{natarajan2021repulsion} becomes inefficient and too slow to be usable.
For this study, we therefore implemented the MCMC in conjunction with the chaperones algorithm \cite{Betancourt2016FlexibleMF,betancourt2020random}, a further development of split-merge Markov Chain algorithms \cite{jain2004split} targeted at applications with many small clusters. As initialization of the chain we use a $k$-medoids clustering from the provided distances. 
The MCMC estimates a co-clustering matrix containing the probabilities of coin image pairs to be struck by the same die, on the basis of which a cluster-label point-estimate is derived using the SALSO algorithm~\cite{dahl2021search}.

\subsection{Comparison with Difference-of-Gaussian keypoints}\label{subsec:CADS}

\begin{figure}[!t]
     \centering
     \begin{subfigure}[b]{0.49\columnwidth}
         \centering
         \includegraphics[width=\textwidth]{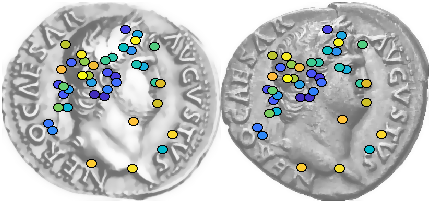}
         \caption*{GP, same-die.}
     \end{subfigure}
     \begin{subfigure}[b]{0.49\columnwidth}
         \centering
         \includegraphics[width=\textwidth]{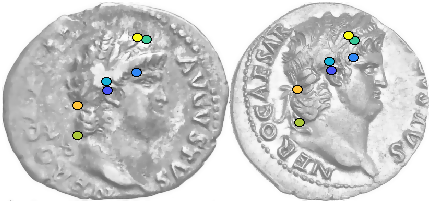}
         \caption*{GP, different-die.}
     \end{subfigure}

     \begin{subfigure}[b]{0.49\columnwidth}
         \centering
         \includegraphics[width=\textwidth]{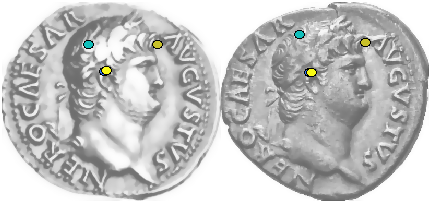}
         \caption*{ORB, same-die.}
     \end{subfigure}
     \begin{subfigure}[b]{0.49\columnwidth}
         \centering
         \includegraphics[width=\textwidth]{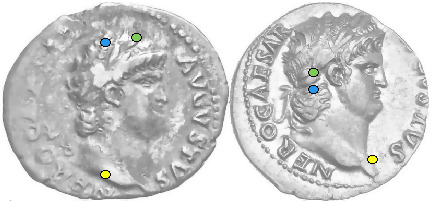}
         \caption*{ORB, different-die.}
     \end{subfigure}
\caption{Illustration of keypoints after low-distortion matching (indicated by colors) for same-die and different-die obverse pairs.
For the same circular regions, 190 candidate GP-keypoints, respectively the strongest 190 ORB-keypoints, were provided for matching. 
We observed that providing more ORB-keypoints resulted in more matches among same-die pairs, but did not improve semantic consistency of matches between different-die pairs.
} 
\label{Fig:matchComparison}
\end{figure}

The CADS software \cite{taylor2020computer} generates dendograms in an unsupervised manner, to support manual hierarchical clustering of coin images for die analysis. By examining the branches of the 
dendogram, researchers have to decide whether coins with shared features were struck from the same die. In this sense, CADS is a pre-sorting tool suggesting groups of coins with a higher chance to belong to the same die, rather than a method for largely automating the process of die analysis.
Moreover, in contrast to our method, CADS derives similarity scores for image pairs via off-the-shelf keypoint extraction tools, utilizing ORB~\cite{rublee2011orb}, a further development of the Scale Invariant Feature Transform~\cite{tareen2018comparative}.
In analogy to our pre-matching step, CADS uses the Hamming distances of keypoint descriptor vectors to match the ORB-keypoints for each image pair and to calculate a similarity score by averaging the Hamming distances of up to twenty strongest matches. Yet, ORB and related Difference-of-Gaussian based methods are not specifically designed for die analysis, but rather for applications such as object recognition, tracking or registration. 
Consequently, ORB generates many close-by keypoints that refer to a feature at different scales, for instance corner points or centres of blobs of an object to be detected in photographs taken from greatly varying distances or angles, and does not place keypoints at coarse arcs or line segments. By contrast, the GP detects spacially balanced keypoints, and avoids keypoints that would be redundant in a die study application (Fig.~\ref{Fig:GPvsORB}). As a result,
GP-keypoint positions are perceptually consistent with the relevant visual content of the images, and the number of matched keypoints can provide valuable information for the dissimilarity assessment. Fig.~\ref{Fig:matchComparison} illustrates low-distortion matchings for same- and different-die coin pairs from GP, respectively ORB, candidate keypoints. 
While there are alternative methods for keypoint detection, e.g., based on wavelets \cite{fauqueur2006multiscale}, the GP-approach has the advantage of spatially balancing keypoint positions using the amount of additional information they provide.

\begin{figure*}[!t]
     \centering
     \begin{subfigure}[b]{0.32\textwidth}\centering
        \includegraphics[width=0.49\columnwidth]{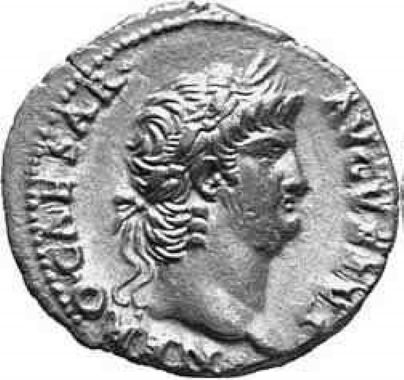}
                \hspace{0.48\columnwidth}

        \includegraphics[width=0.49\columnwidth]{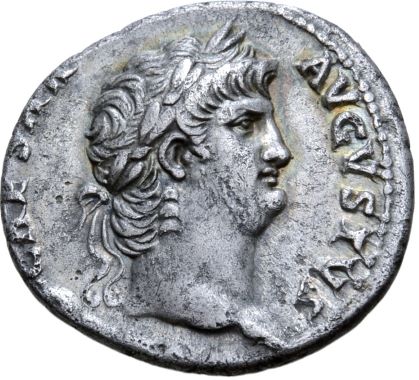}
        \includegraphics[width=0.49\columnwidth]{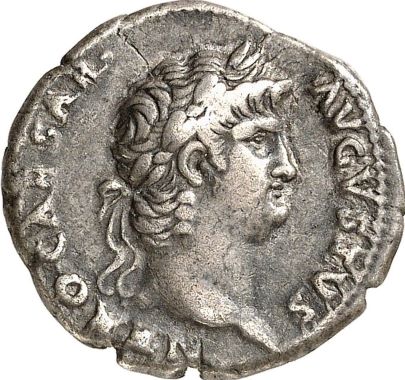}

        \includegraphics[width=0.49\columnwidth]{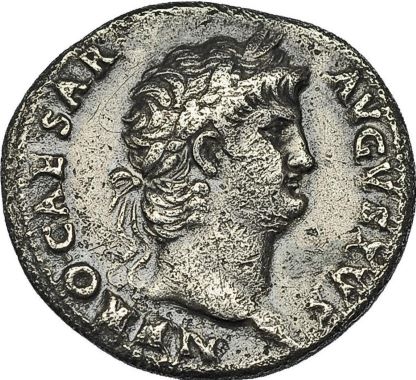}
        \includegraphics[width=0.49\columnwidth]{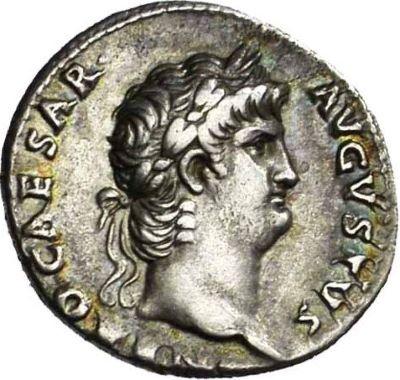}
        \caption{Die 1}
     \end{subfigure}
     \rulesep
     \begin{subfigure}[b]{0.15\textwidth}\centering
        \includegraphics[width=\columnwidth]{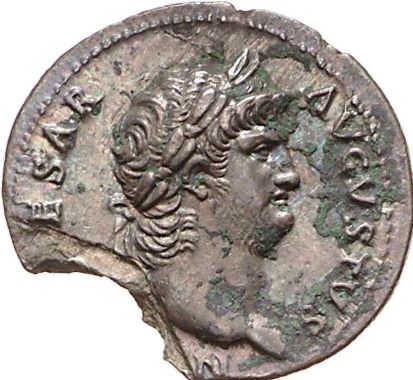}
        
        \includegraphics[width=\columnwidth]{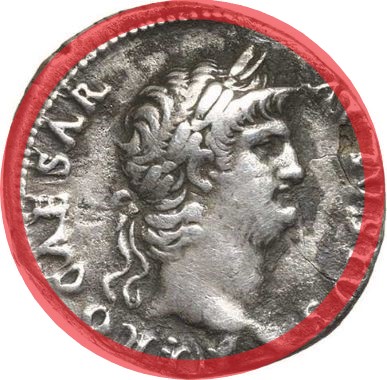}
                \caption{Die 2}
     \end{subfigure}
     \rulesep
     \begin{subfigure}[b]{0.49\textwidth}\centering
        \includegraphics[width=0.32\columnwidth]{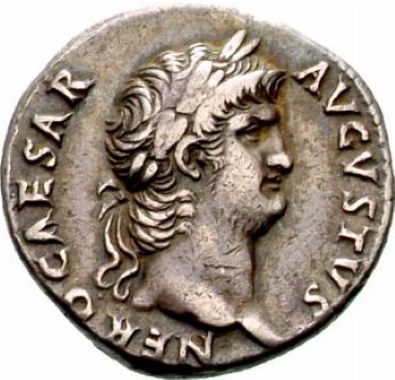}
        \includegraphics[width=0.32\columnwidth]{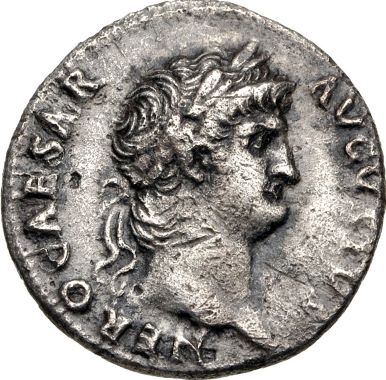}
        \includegraphics[width=0.32\columnwidth]{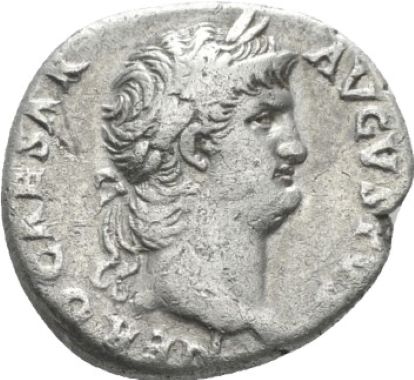}
        
        \includegraphics[width=0.32\columnwidth]{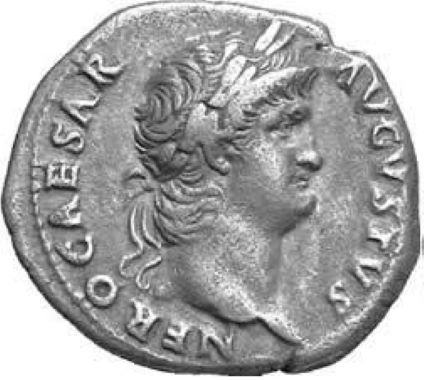}
        \includegraphics[width=0.32\columnwidth]{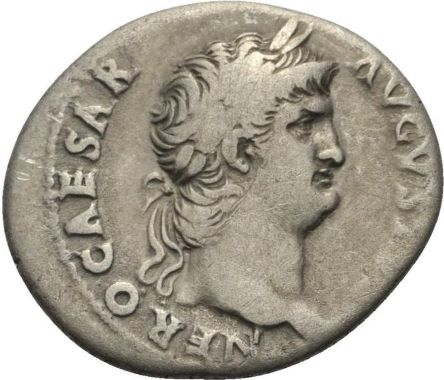}
        \includegraphics[width=0.32\columnwidth]{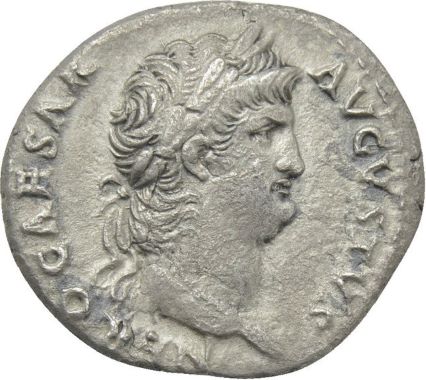}
        
        \includegraphics[width=0.32\columnwidth]{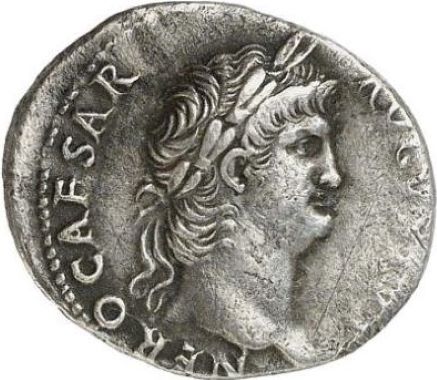}  
        \includegraphics[width=0.32\columnwidth]{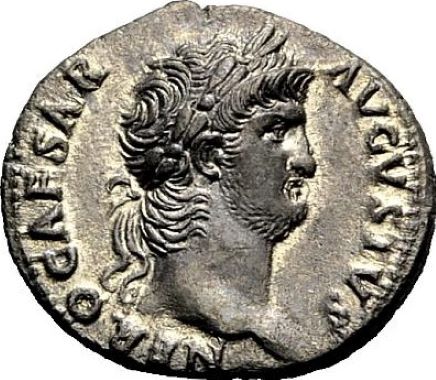}
        \includegraphics[width=0.32\columnwidth]{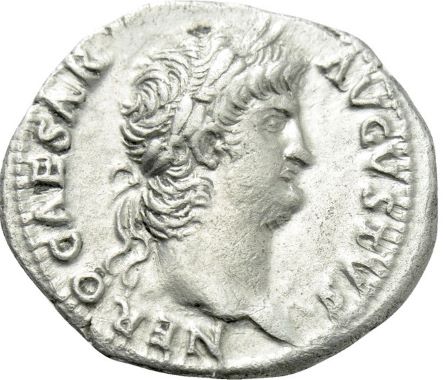}
        \caption{Die 3}
        \label{Fig:DieSquare}
     \end{subfigure}
\caption{Prediction result for obverses struck from three distinct dies. The dies only differ in minute details, and were likely made by the same engraver. Still, the model separated all images correctly, and only the circled image was not clustered together with the first coin struck from Die 2 (sensitivity $0.5$). None of the images was erroneously clustered together with any images of the remaining 294 disjoint classes. 
}
\label{Fig:PredExpl2}
\end{figure*} 
\section{Results}\label{Sec:Results}

We demonstrate the performance of our method through an analysis of 2\,866 images, representing 1\,135 individual Neronian denarii. 
In Sec.~\ref{Sec:DieAnalysisDiscussion}, we further describe the visual validation to establish the true classes against which we compare the clustering results of our method. This visual validation process coincides with conducting a die study from the computational cluster predictions, and in Sec.~\ref{Sec:DieAnalysisDiscussion} we indicate the manual work saved in comparison to conducting a brute-force visual die analysis without computational assistance.

As a global accuracy metric we report in Table~\ref{table:summary} the Normalized Mutual Information (NMI)~\cite{Strehl2003NMI}, 
which quantifies the reduction in the entropy of class labels $l$, conditional on knowing the predicted cluster labels $c$: 
\[
\mathrm{NMI}(l,c) = \frac{I(l,c)}{\sqrt{H(l)H(c)}},
\]
where $H$ denotes the entropy, $I(l,c)=H(l)-H(l\mid c)$ the mutual information, and $H(l\mid c)$ the conditional entropy of die labels within each cluster. The range of the NMI is $[0,1]$, with larger values corresponding to higher accuracy.

\begin{figure*}[!t]
\centering
\begin{subfigure}[b]{0.49\textwidth} 
        \includegraphics[width=0.24\textwidth]{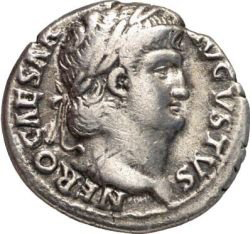}
        \includegraphics[width=0.24\textwidth]{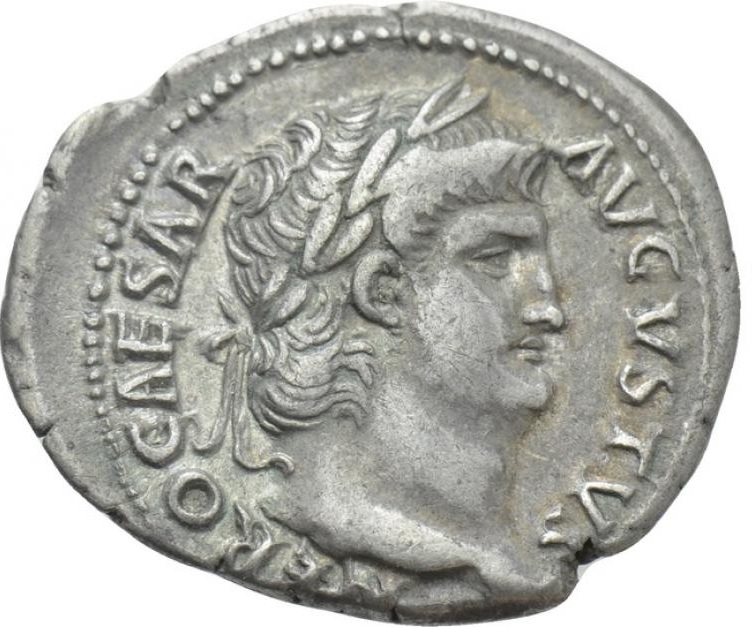}
        \includegraphics[width=0.24\textwidth]{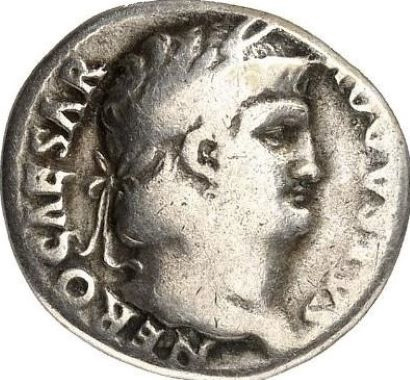}

        \includegraphics[width=0.24\textwidth]{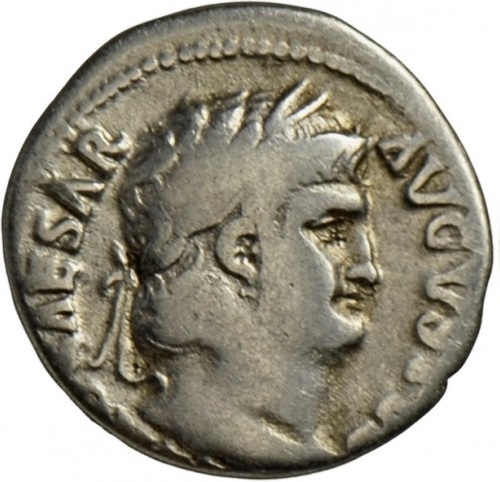}
        \includegraphics[width=0.24\textwidth]{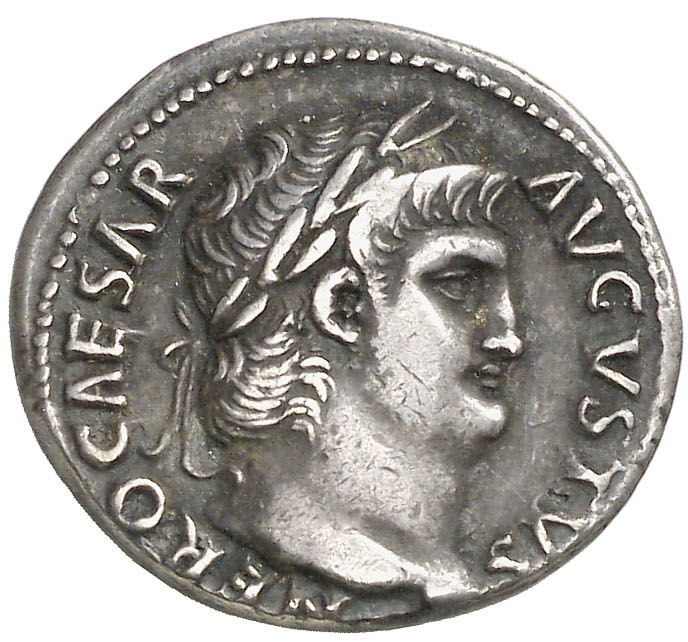}
        \includegraphics[width=0.24\textwidth]{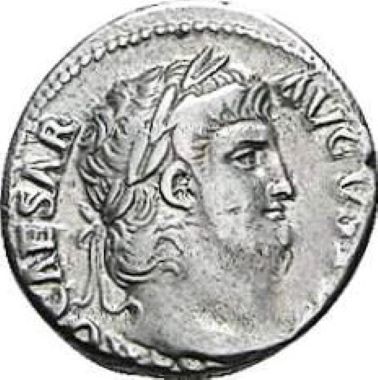}
        \includegraphics[width=0.24\textwidth]{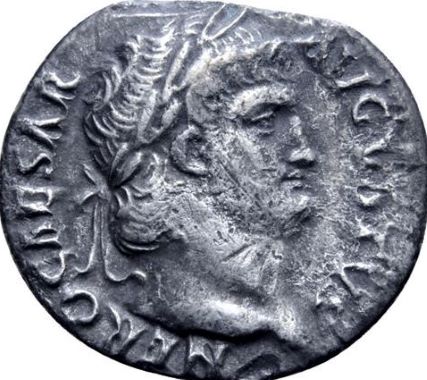}
        
        \includegraphics[width=0.24\textwidth]{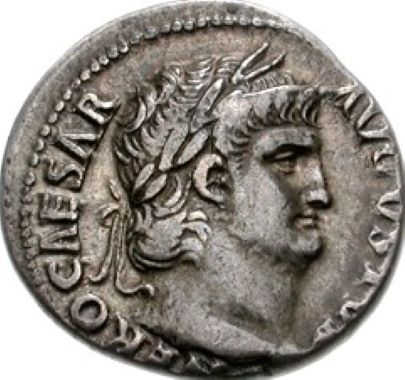}
        \includegraphics[width=0.24\textwidth]{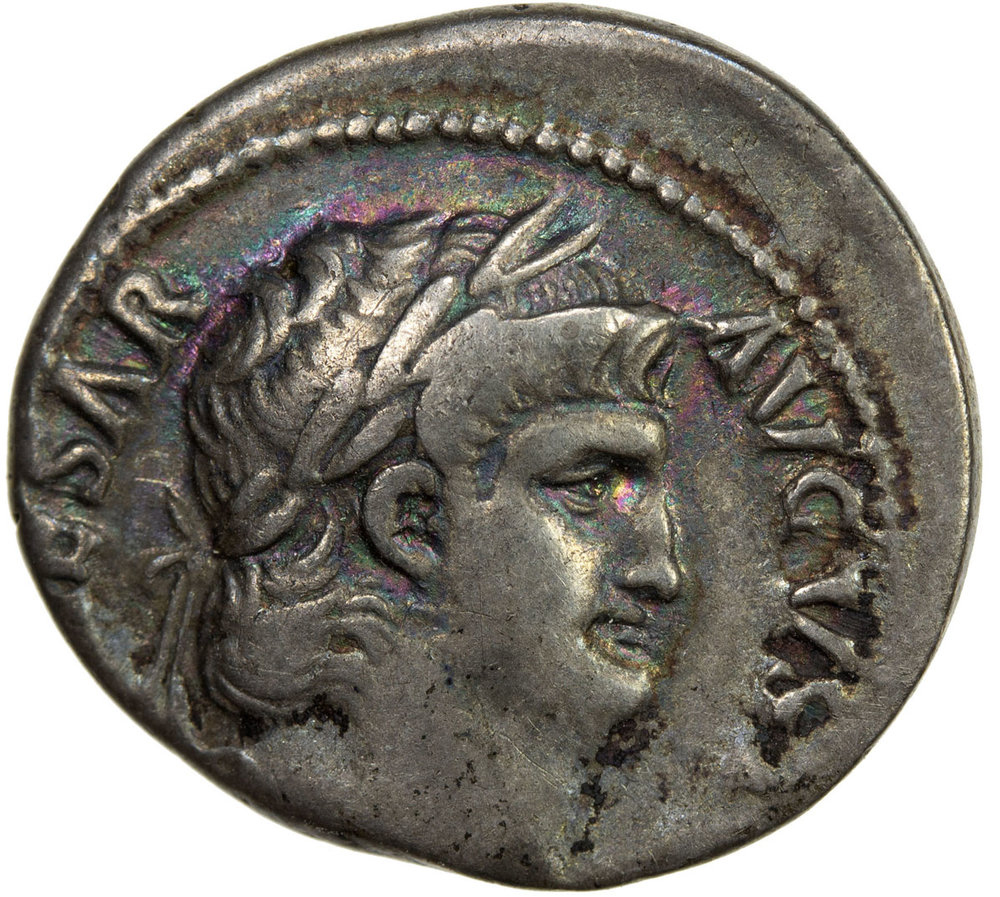}
        \includegraphics[width=0.24\textwidth]{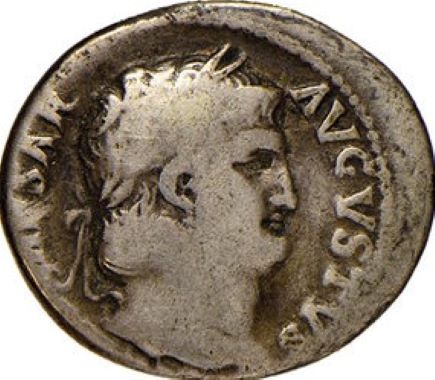}
        \includegraphics[width=0.24\textwidth]{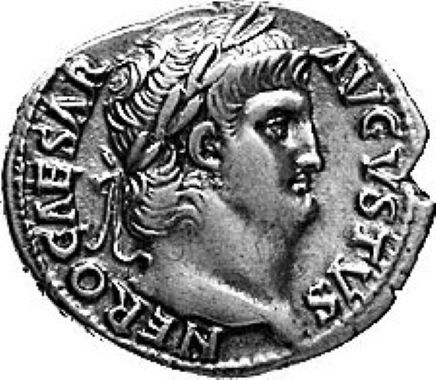}
\caption{Die $\circ$} 
\end{subfigure}
\begin{subfigure}[b]{0.5\textwidth} 
\centering
  \includegraphics[width=\textwidth]{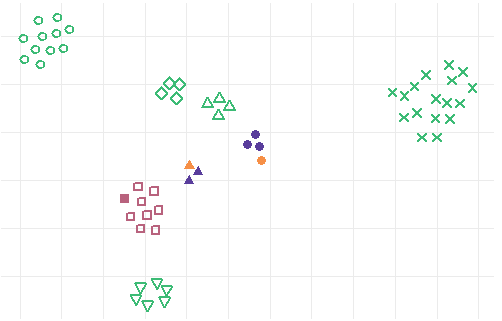}
\caption{GP-distance based $t$-SNE illustration for the displayed coins.} 
\end{subfigure}
%
\hfill
\begin{subfigure}[b]{0.24\textwidth}
    \centering
    \includegraphics[width=0.48\textwidth]{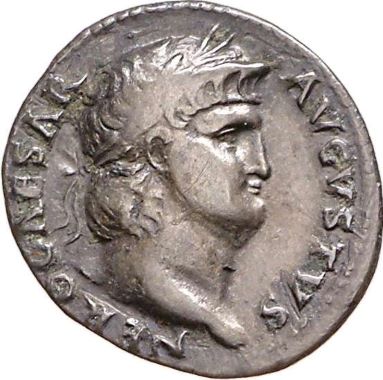}      
    \includegraphics[width=0.48\textwidth]{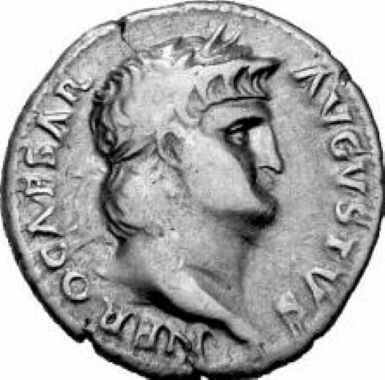}      
    
    \includegraphics[width=0.48\textwidth]{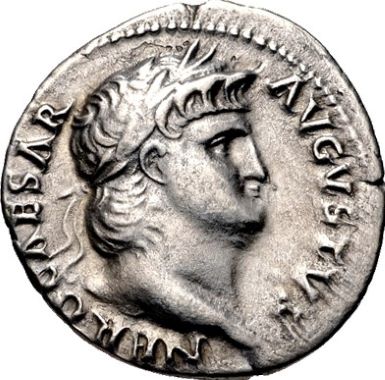}       
    \includegraphics[width=0.48\textwidth]{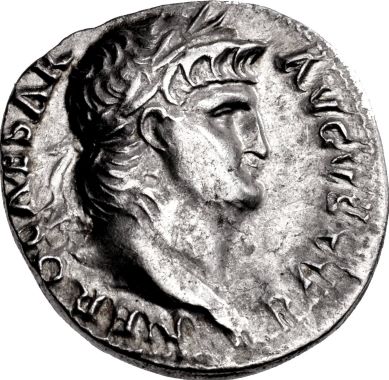}
    \caption{Die $\Diamond$}  
\end{subfigure}
\rulesep
\begin{subfigure}[b]{0.24\textwidth}
    \centering
    \includegraphics[width=0.48\textwidth]{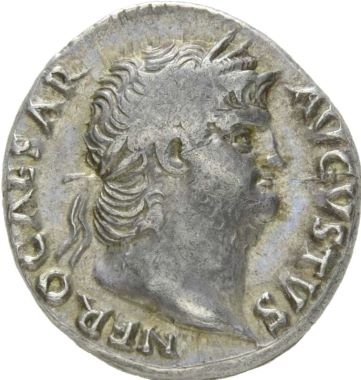}      
    \includegraphics[width=0.48\textwidth]{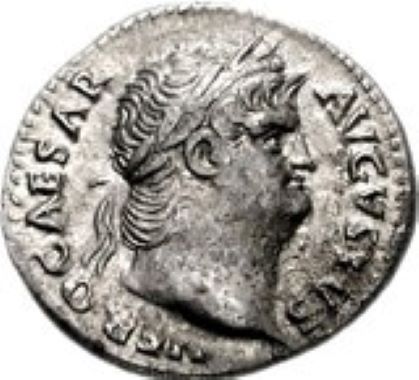}       
    
    \includegraphics[width=0.48\textwidth]{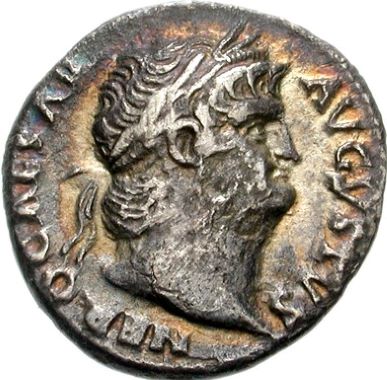}       
    \includegraphics[width=0.48\textwidth]{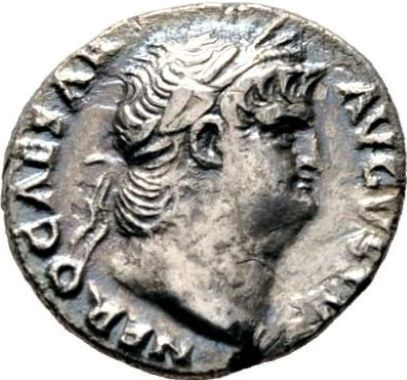}  
\caption{Die $\vartriangle$}  
\end{subfigure}
\rulesep
\begin{subfigure}[b]{0.24\textwidth}
    \includegraphics[width=0.48\textwidth]{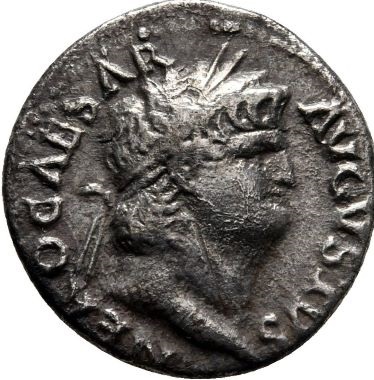}  
    
    \includegraphics[width=0.48\textwidth]{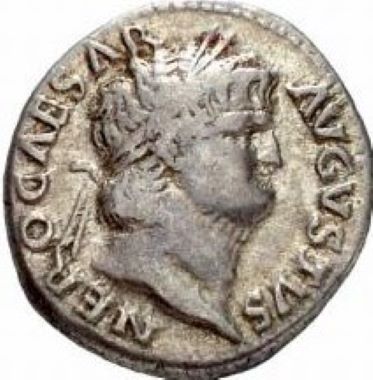}     
    \includegraphics[width=0.48\textwidth]{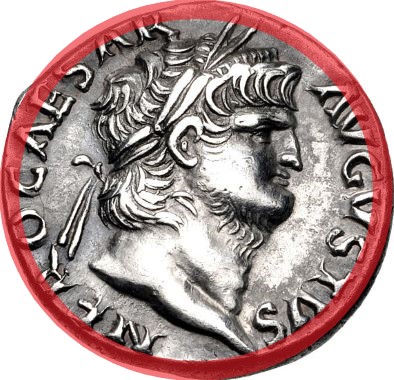}
    \caption{Die $\blacktriangle$}  
\end{subfigure}
\rulesep
\begin{subfigure}[b]{0.24\textwidth}
    \centering
    \includegraphics[width=0.48\textwidth]{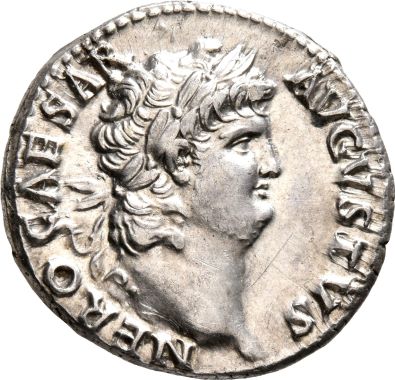}       
    \includegraphics[width=0.48\textwidth]{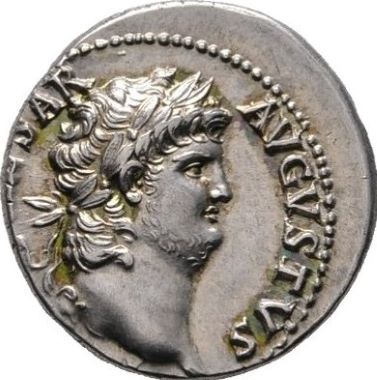}       
    
    \includegraphics[width=0.48\textwidth]{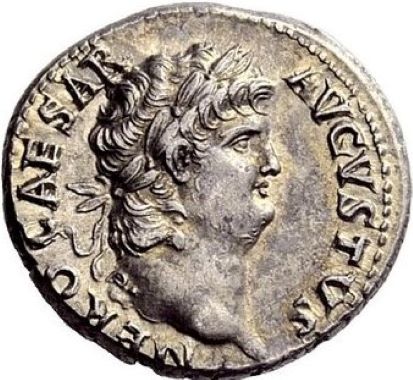}     
    \includegraphics[width=0.48\textwidth]{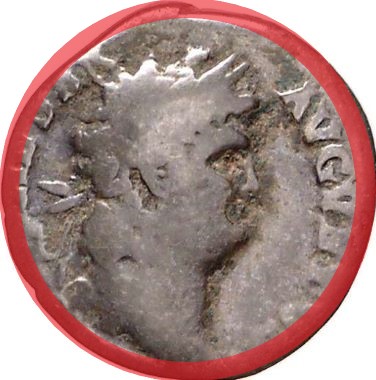}  
    \caption{Die $\bullet$}  
\label{Fig:mixf}
\end{subfigure}
\hfill
     \begin{subfigure}[b]{0.37\textwidth}
    \centering
    \includegraphics[width=0.32\textwidth]{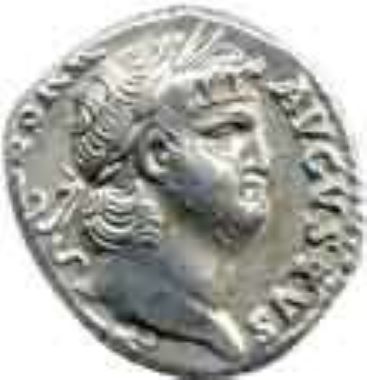}
    \includegraphics[width=0.32\textwidth]{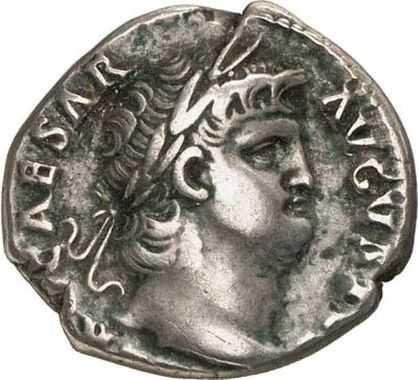}  
    \includegraphics[width=0.32\textwidth]{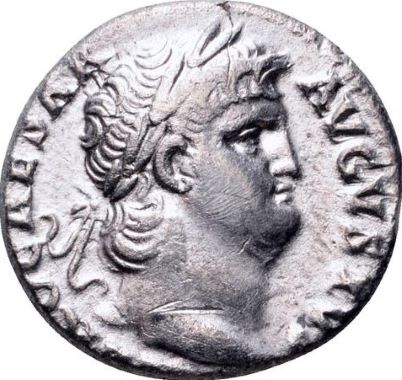}  
    
    \includegraphics[width=0.32\textwidth]{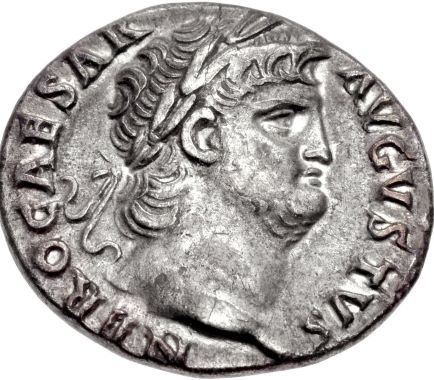}  
    \includegraphics[width=0.32\textwidth]{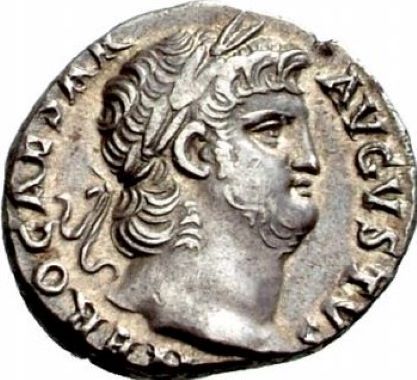}  
    \includegraphics[width=0.32\textwidth]{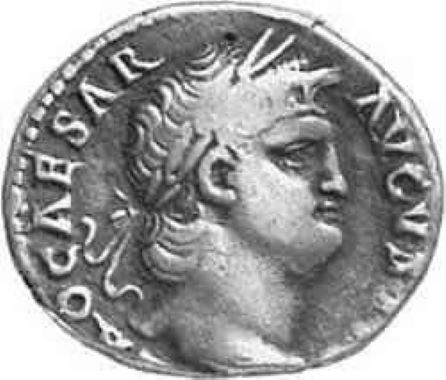}
    \caption{Die $\triangledown$}  
\end{subfigure}
\rulesep
\begin{subfigure}[b]{0.49\textwidth}
    \centering
    \includegraphics[width=0.23\textwidth]{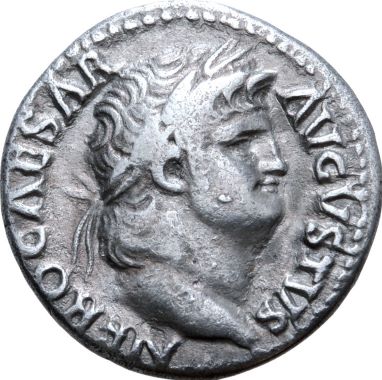}
    \includegraphics[width=0.23\textwidth]{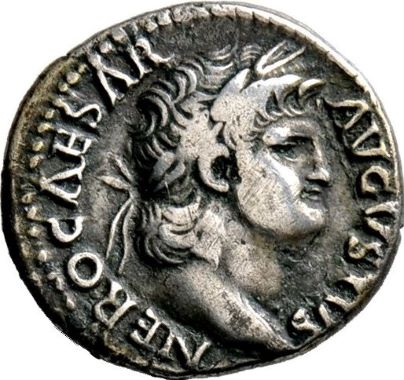}
    \includegraphics[width=0.23\textwidth]{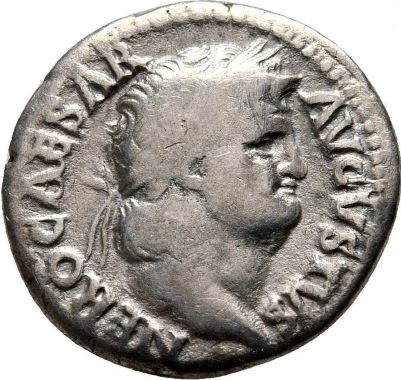}       
    \includegraphics[width=0.23\textwidth]{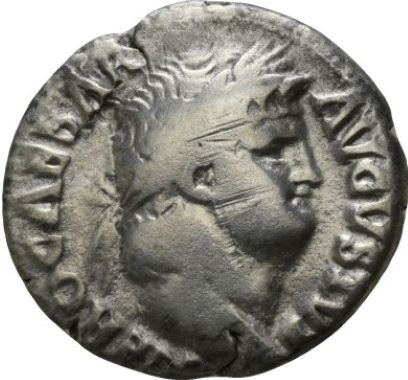}

    \includegraphics[width=0.23\textwidth]{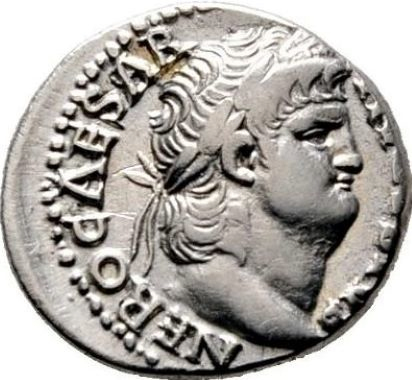}   
    \includegraphics[width=0.23\textwidth]{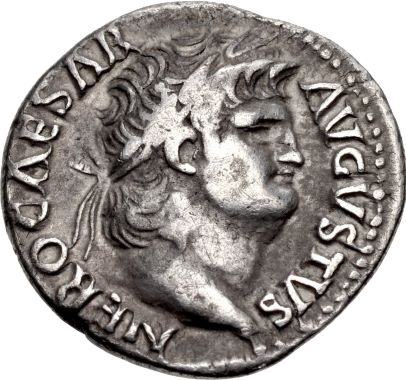}
    \includegraphics[width=0.23\textwidth]{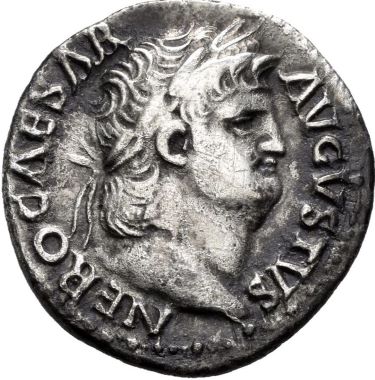}
    \includegraphics[width=0.23\textwidth]{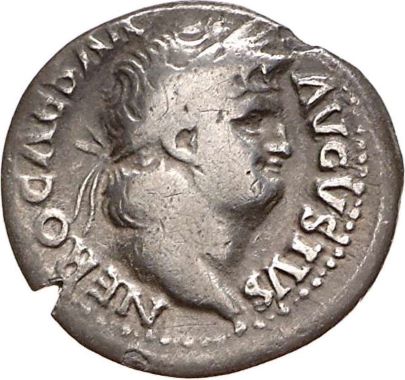}
    \caption{Die $\square$}  
\label{Fig:DieBlacktriangle}
\end{subfigure}
\rulesep
\begin{subfigure}[b]{0.11\textwidth}
    \centering
    \includegraphics[width=\textwidth]{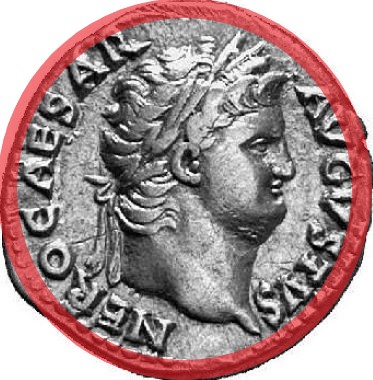}
    \caption{Die $\blacksquare$}  
\end{subfigure}
\hfill
\caption{Coin faces struck by eight of the 297 dies of the obverse dataset: For the dies $\circ, \Diamond, \vartriangle$ and  $\triangledown$, cluster predictions coincide with the true dies. For the die $\blacktriangle$ our method identified correctly two out of the three coins (sensitivity $0.67$), for die $\bullet$ three out of four coins were identified (sensitivity $0.75$). The coin of the singleton die $\blacksquare$ was incorrectly assigned the same cluster label as the coins of die $\square$, contributing to the $\mathrm{FDR}$ of both die $\square$ ($1/9\approx 0.11$) and $\blacksquare$ ($8/9\approx 0.89$). The displayed images were not clustered together with any images of the remaining 289 dies. 
Subfigure (b) shows a two-dimensional representation of the displayed coins, generated from the GP-keypoint dissimilarities by multi-dimensional scaling and $t$-SNE \cite{BronsteinMDS,Hinton_tSNE}. Additionally, the coins shown in Fig.~\ref{Fig:PredExpl2} are indicated by $\times$. 
}
\label{Fig:PredExplMixes}
\end{figure*}

\begin{figure*}[!t]
\centering
\begin{subfigure}[b]{\textwidth}
    \includegraphics[width=\textwidth]{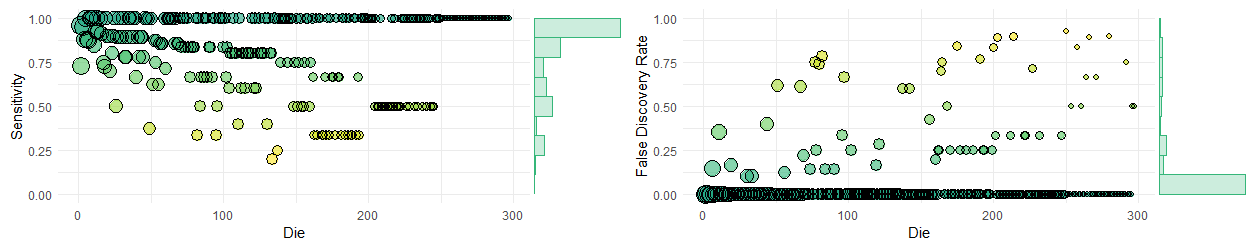}
    \caption{Obverses: 297 dies.}
    \label{Fig:Obv}
\end{subfigure}
\begin{subfigure}[b]{\textwidth}
    \includegraphics[width=\textwidth]{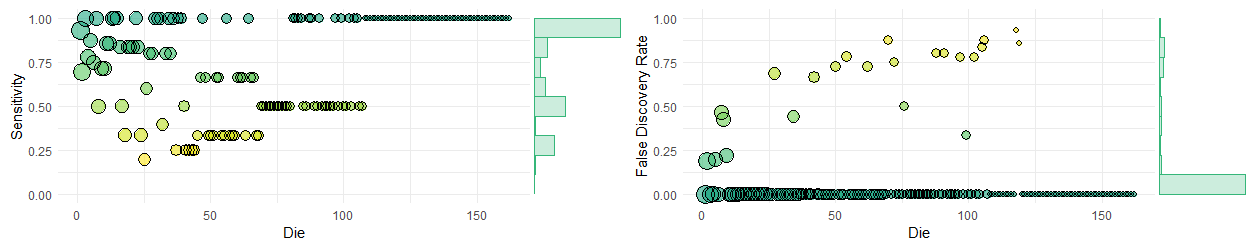}
    \caption{Reverses \emph{RIC} I\textsuperscript{2} 60: 162 dies.}
    \label{Fig:Rev60}
\end{subfigure}
\caption{Sensitivities (optimal value one) and false discovery rates (optimal value zero) for all dies in the obverse and the largest reverse dataset. Right margin histograms summarize the frequencies of dies in the respective ranges. Bullet sizes represent class sizes, i.e., the number of images with which a die is represented in the dataset.}
\label{Fig:FDR}
\end{figure*}

\begin{figure*}[!ht]
\caption{
Sensitivities (optimal value one) and false discovery rates (optimal value zero) for all dies of the datasets \emph{RIC} I\textsuperscript{2} 53; 55; 57; 62. Right margins summarize the frequencies of dies in the respective ranges. Bullet sizes represent class sizes, i.e., the number of images with which a die is represented in the dataset.
}
\label{Fig:FDRrest}
\subcaptionbox*{Reverses \emph{RIC} I\textsuperscript{2} 53}{\includegraphics[width=\textwidth]{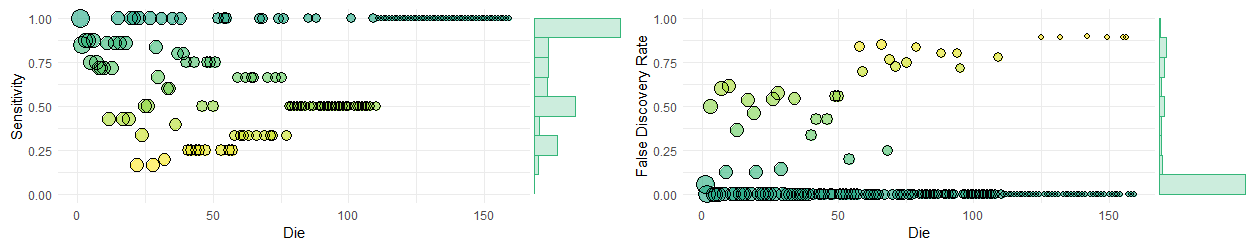}}

 \vspace{2em}
\subcaptionbox*{Reverses \emph{RIC} I\textsuperscript{2} 55}{\includegraphics[width=0.49\textwidth]{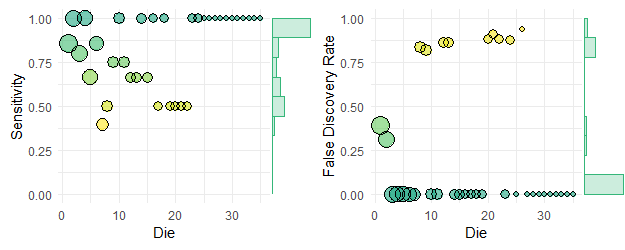}}
\subcaptionbox*{Reverses \emph{RIC} I\textsuperscript{2} 57}{\includegraphics[width=0.49\textwidth]{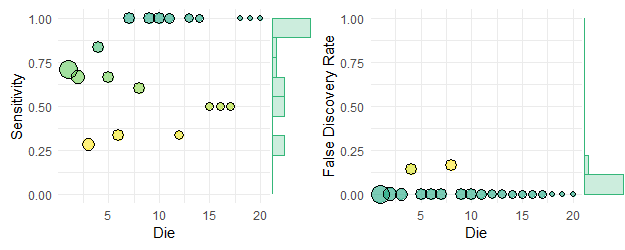}}

 \vspace{2em}
\subcaptionbox*{Reverses \emph{RIC} I\textsuperscript{2} 62}{\includegraphics[width=\textwidth]{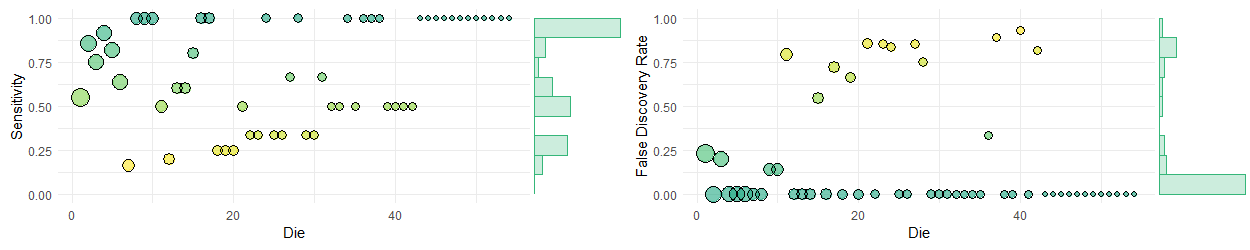}}
\end{figure*}

We further assess the success of our model by considering the following two statistics for each true class (precise definitions further below).
\begin{itemize}
    \item \emph{Sensitivity}, i.e., the proportion of images from the class that are correctly identified, and
    \item \emph{False discovery rate} ($\textrm{FDR}$), quantifying erroneous ``mixing'' with coins from different classes.
\end{itemize} 
Our method was able to predict a large number of entirely correct clusters, even when they included images of differing acquisition characteristics or of badly preserved coins. However, it failed to allocate some images to their true class -- resulting in lower sensitivity -- or allocated an image of a disjoint class to the images of a cluster -- resulting in a nonzero $\textrm{FDR}$ and lower sensitivity. For examples of such cases see Figs.~\ref{Fig:PredExpl2} and~\ref{Fig:PredExplMixes}, in which duplicate images have been removed for the visualization.  
We define the sensitivity for a class as the ratio of the largest intersection of any predicted cluster with the class, to the size of the class. 
We define the false discovery rate for a class as the ratio of the number of images that were incorrectly assigned the same cluster label as any image from the class, to the sum of class size and number of incorrectly attributed images. Note that clusters containing images from different classes contribute to the $\textrm{FDR}$ of each of the classes involved.
The sensitivities and false discovery rates for every obverse die, and every reverse die of one large reverse dataset, are illustrated in Fig.~\ref{Fig:FDR} (see Fig.~\ref{Fig:FDRrest} for remaining reverse types). Table~\ref{table:summary} summarizes these statistics, reporting for each dataset averages over all classes weighted by cluster size.

\begin{figure}[htb] 
 \begin{subfigure}[b]{0.24\columnwidth}
        \centering
        \includegraphics[width=\columnwidth]{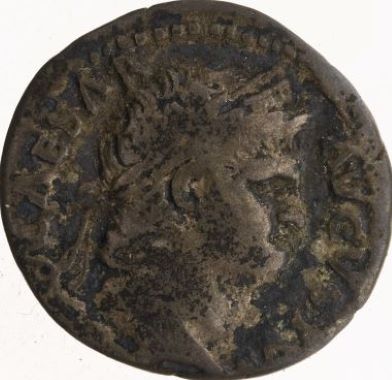}
        
        \includegraphics[width=\columnwidth]{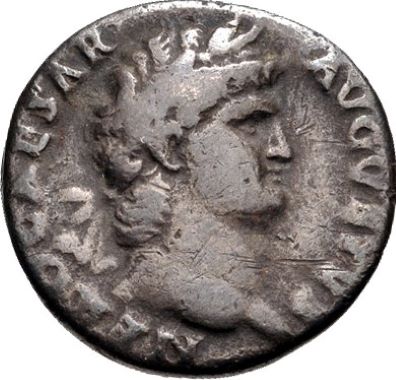}
        \caption*{\emph{good}}
 \end{subfigure}
  \begin{subfigure}[b]{0.24\columnwidth}
        \centering
        \includegraphics[width=\columnwidth]{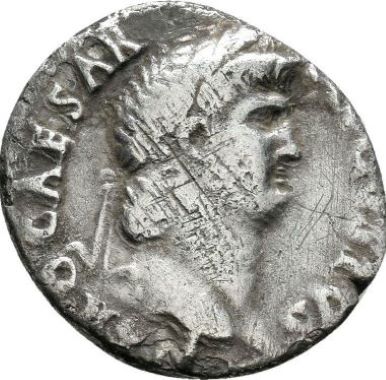}
        
        \includegraphics[width=\columnwidth]{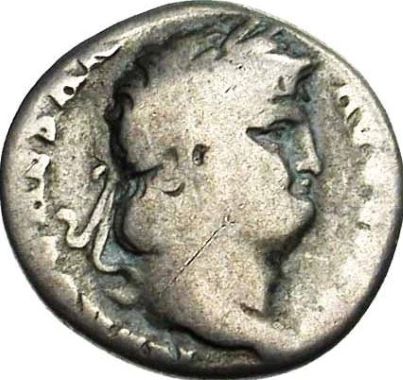}
        \caption*{\emph{very good}}
 \end{subfigure}
  \begin{subfigure}[b]{0.24\columnwidth}
        \centering
        \includegraphics[width=\columnwidth]{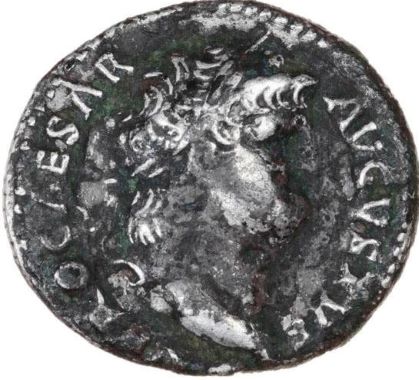}
        
        \includegraphics[width=\columnwidth]{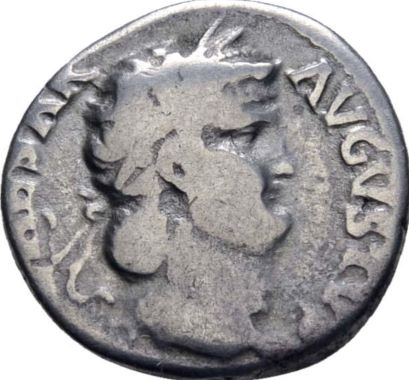}
        \caption*{\emph{fine}}
 \end{subfigure}
  \begin{subfigure}[b]{0.24\columnwidth}
        \centering
                \includegraphics[width=\columnwidth]{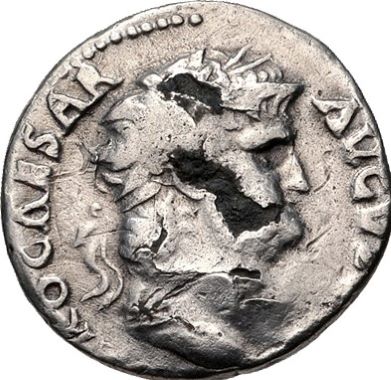}
                
        \includegraphics[width=\columnwidth]{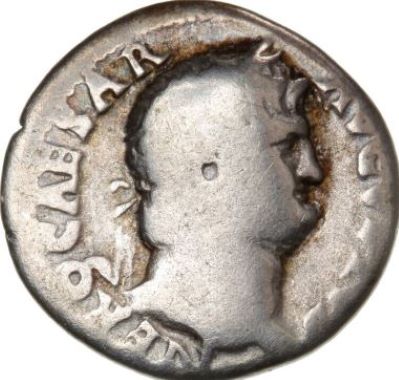}
        \caption*{\emph{very fine}}
 \end{subfigure}
\caption{Examples of badly preserved coins: Images of coins of low preservation grades \emph{good, very good} and \emph{fine} comprise $26\%$ of the sample. \emph{Very fine} coins may still show considerable wear and damage.
}
\label{Fig:badcoins}
\end{figure}

Overall results in Table~\ref{table:summary} show high sensitivity and low $\mathrm{FDR}$. Yet, there are differences between the six separately studied numismatic samples. Coin faces with more details exhibit more discriminative information and yield better results. The highest NMI is achieved for the obverses, even though this sample contained substantially more images than any of the reverse datasets. The Vesta temple of \emph{RIC} I\textsuperscript{2} 62 emerged as the most challenging motif, because it contains the least amount of details. Performance was relatively little affected by removing duplicate images ($20.8\%$) from the obverse dataset, pairs of which are easier to detect than general pairs from the same class, and by removing coins of the three lowest preservation grades (see Fig.~\ref{Fig:badcoins}).

For the reported obverses results, dissimilarities were derived from 300 GP-keypoints per image (190 keypoints and smaller circular region led to only slightly weaker clustering results). We determined the number of keypoints by visual inspection on a well-preserved coin, selecting a total number before the GP starts placing keypoints at semantically meaningless positions for reasons of querying remaining large regions for uncertainty reduction. We ran 750\,000 MCMC iterations (half of them burn-in, thinning to every tenth iteration). As hyper-parameter settings we chose a mean cluster size of five and variance of 15 (see Table~\ref{tab:sen} for comments on hyper-parameter sensitivity). To target lower FDR over higher sensitivity, and given the high percentage of poorly preserved sample coins, we further implemented a hard constraint in the chaperones algorithm, preventing at each iteration step label re-assignments to clusters with current size exceeding $25$.

\begin{table*}[!t]
\caption{Performance summary: For each dataset we indicate the number of images and classes/dies, as well as the percentage of images of the three lowest preservation grades, \emph{good, very good} and \emph{fine} (lp). As performance metrics for our method, we report the normalized mutual information (NMI), and an average of the sensitivities and false discovery rates (FDR) over all dies of the sample. Besides reporting our results on the full obverses dataset (Obv) we also tested our method on the dataset after removing duplicate images of the same coin (Obv\textsuperscript{-du}), and after removing duplicate images as well as the images of the three lowest preservation grades (Obv\textsuperscript{-du-lp}). 
}
\label{table:summary}
\centering
\ra{1.2}
\begin{tabular}[t]{@{}lccccccccc@{}}\toprule 
&\multicolumn{3}{c}{\makecell{Obverses \emph{RIC} I\textsuperscript{2} 53; 55; 57; 60; 62}}&\multicolumn{5}{c}{\makecell{Reverses \emph{RIC} I\textsuperscript{2}}}\\
\cmidrule(lr){2-4}\cmidrule(lr){5-9}
 Dataset   & Obv & Obv\textsuperscript{-du} & Obv\textsuperscript{-du-lp} &  53 & 55 & 57     & 60       &  62          \\
\midrule
$\#$ Images        & $1434$    & $1135$    & $812$  &  $512$   & $122$    & $100$    & $478$    &  $220$       \\
$\#$ Classes/dies  &   $297$   &   $297$   &  $269$ & $159$    & $35$      & $20$      & $162$    &  $54$        \\
lp   &   $26\%$  &   $28\%$  &  --    & $25\%$   & $30\%$    & $20\%$    & $28\%$   &  $22\%$      \\
\midrule
FDR                & $0.067$  & $0.107$   & $0.019$ & $0.143$ & $0.235$    & $0.017$   & $0.093$  & $0.189$      \\
Sensitivity        & $0.840$  & $0.795$   & $0.712$ & $0.695$ & $0.811$    & $0.710$   & $0.736$  & $0.682$      \\
NMI     & $\mathbf{0.959}$ & $\mathbf{0.947}$& $\mathbf{0.949}$& $\mathbf{0.921}$& $\mathbf{0.883}$& $\mathbf{0.895}$& $\mathbf{0.936}$& $\mathbf{0.876}$ \\
\bottomrule
\end{tabular}
\end{table*}%


\section{Discussion}\label{Sec:discussion}
\subsection{Implications for large scale die studies}\label{Sec:DieAnalysisDiscussion}

The method presented in this article can reduce the time investment for large die studies by several orders of magnitude. To illustrate the dramatic reduction using the computational cluster predictions, we outline in this section the visual verification process we conducted to establish the true die classification for the obverse sample (see App.\ B for full details for all obverses and reverses). 

A brute-force visual die analysis of the 1\,434 obverse images would require more than a million pairwise comparisons. With an accuracy of 0.959 (NMI), the number of comparisons required to validate the model's prediction results and to modify them where necessary, was reduced to a fraction of this enormous amount. Validation and modification was done in the following two steps. 
First, to correct false discoveries, each estimated cluster with more than one image was inspected to identify potential images belonging to different classes that had been erroneously clustered together. Breaking-up such predicted clusters yielded a total number of $\widetilde{K}=505$
clusters (including singleton clusters), each now containing only images of the same class. Second, to correct sensitivity imperfections, a representative image of a preferably well preserved coin was chosen from each of the (non-singleton) clusters, and all representatives -- including the images of the singleton clusters -- were compared to each other to merge clusters belonging to the same class that had been missed by our method. With $K$ denoting the true number of classes, a simple combinatorial argument implies that the process of validation and potential merges requires at most $2K\widetilde{K}-K^2-\widetilde{K}^2/2-\widetilde{K}/2$
comparisons to identify the $\widetilde{K}-K$ pairs that belong to the same class. In our sample, with $K=297$ and thus at most 84\,000 comparisons, this corresponds to a reduction of at least $\mathbf{91.8\%}$ from the number of comparisons one is confronted with in order to assess the 1\,436 images directly.  This bound comes close to the number of comparisons necessary for a visual verification of a perfect oracle prediction, which requires $K(K-1)/2$ comparisons (ca.\ 44\,000 for our obverse sample) and corresponds to a reduction of $95.7\%$ from a brute-force visual die analysis. The exact reduction amount depends on the particular order in which the representative coins are considered, influencing when clusters are merged. We recommend to start visual verification with the badly preserved representatives, and to compare each to the well preserved representatives first. Since the majority of coins contributing to sub-optimal sensitivity are badly preserved, this strategy requires fewer comparisons than the worst-case bound given above.

\begin{figure}[!t]
\centering
\includegraphics[width=\columnwidth]{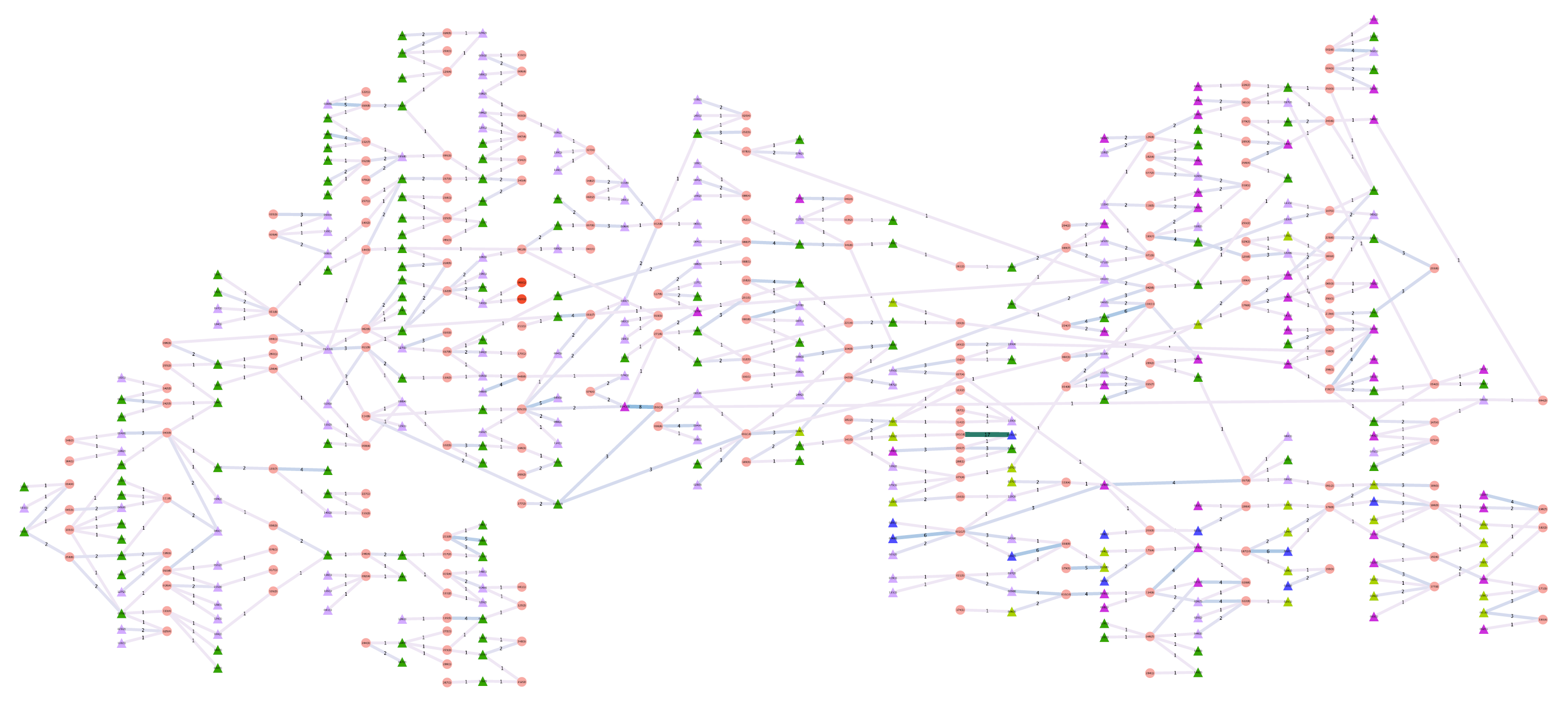}
\caption{Die combination diagram linking ca.\,$70\%$ of the dies studied in this article: Vertices represent obverse dies (circles) and reverse dies (triangles, colored by numismatic type). An obverse and reverse die are connected by an edge if they have been used to strike a coin in the sample. The structure of large die combination diagrams can provide new insights into questions of chronology, die lifetimes, or mint organization.}
\label{Fig:dld}
\end{figure}

\begin{figure}[htb]
\centering
\includegraphics[width=\columnwidth]{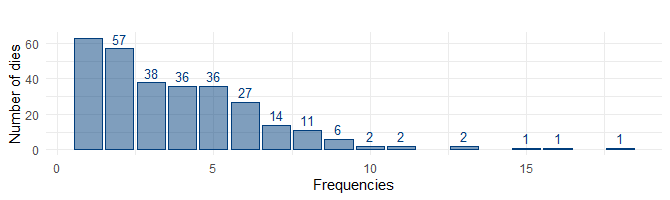} 
\caption{Frequency chart for the obverse dataset:
Estimates for the total number of dies used to mint a series of ancient coins can be based on the number of dies with respective coin frequencies observed in the sample. 
The predicted total number of dies used to mint an issue can serve as a proxy to estimate the total number of minted coins \cite{esty1986estimation,esty2011geometric}. 
With the ability to conduct die studies of previously unattainable sizes, it will be interesting to compare numismatic practices for coinage size estimation with related methodology from other disciplines, such as microbial diversity estimation in the life sciences~\cite{bunge2014estimating}.}
\label{Fig:freqchart}
\end{figure}

The decrease in required comparisons translates to saving several thousand hours of manual work -- not least because, on average, each of the remaining comparisons is less time-consuming than those in a brute-force die study. Most images are already correctly allocated and many true classes correctly predicted, which means, in practice, that visual validation mostly involves corrections for low-quality images or images of badly preserved coins.
The time reduction allowed for the completion of the validation process within a few days -- enough time to develop a familiarity with the material, which allowed for an analysis of hands, styles, graffiti and other markings. In fact, the model presented here will allow researchers to spend more time on numismatic and art-historical analysis of their material, in addition to providing information about the minting process through die combination diagrams (Fig.~\ref{Fig:dld}), and mint output through statistical analysis of, e.g., frequency charts (Fig.~\ref{Fig:freqchart}).

\subsection{Comparison with Riedones3D}\label{sec:comparison} 
The Riedones3D dataset of Horache et al.~\cite{horache2021riedones3d} comprises of 3D-scans of a hoard find of 1\,035 ancient Celtic coins. The coins were struck from a small number of dies, which is common for hoards of freshly minted coins that did not circulate much. Though the coins suffered from corrosion, they did not suffer from much wear because of their short circulation. As a result, 3D-scans can reveal very fine details. The supervised method introduced in~\cite{horache2021riedones3d} to cluster the 3D-scans by dies requires first a registration step, which, for ancient coins, is a sophisticated problem. Registration is not required by our method, and has not been performed for the results reported in this paper, though registration could improve the performance of our method, in particular for off-centre strikes. In \cite{horache2021riedones3d}, registration is based on deep convolutional autoencoder features \cite{Choy2019FCGF}, trained on 
manually labelled and registered obverse scans.
Pairwise dissimilarities are then derived from the features, before logistic regression is used to assign same-die probabilities for all pairs -- requiring a labelled training set. Clustering is based on a complete graph, with edge weights representing probabilities of adjacent coins to belong to the same die: Predictions are the connected components after thresholding edge weights at a universal value, which has to be learned from labelled examples of the data under consideration.
Riedones3D contains 887 coins of the same obverse type, of which 599 were used for training and validation, while the test set comprised of 288 coins (14 classes). Of the reverse scans, 809 were used for training and validation, and 293 for testing (30 classes). Performances reported on the test sets are an Adjusted Rand Index (ARI) of $0.86$ (reverses), and $0.99$ (obverses). The ARI~\cite{HA85ARI} is the corrected-for-chance percentage of correct decisions made by the method as a binary classifier, here for predicting same-die and different-die pairs. 
For comparison, our method achieved an ARI of $0.83$ on our obverses dataset (
$0.73$ on Obv\textsuperscript{-du}). 
The test sets on which the results in~\cite{horache2021riedones3d} are reported contain relatively few classes -- less than $5\%$ of the number of classes in our obverse dataset -- which can only be expected in exceptional hoard finds. A larger number of classes is likely to exhibit a greater variety of inter-class dissimilarities, 
making the choice of a universal threshold for graph-based clustering challenging. 
The fact that only obverse scans have been used for training in the registration step could be a factor in the performance difference between obverse and reverse test sets in~\cite{horache2021riedones3d}. Moreover, the 200 labelled obverse scans used for registration training -- the preparation of which required two months of work -- consisted of over 2\,000 same-die pairs, which are crucial for training. 
For datasets with a larger number of often smaller sized classes, a substantially greater number of coins would be required to yield a comparable number of same-die pairs. 



\section{Conclusions}\label{Sec:Conclusions}
In this work we introduced a method for automated partitioning of images of ancient coinages by dies used for striking --  an unsupervised clustering problem involving intricate and hard to distinguish images. Our method uses pairwise dissimilarities derived from Gaussian process-based keypoints that enable the use of Bayesian distance microclustering. The re-weighted Gaussian process covariance structure corresponds to a convolution of coin relief edge information for the iterative choice of maximally informative keypoint locations. The performance of our method on a dataset of 2\,868 images of obverses and reverses of 1\,135 Roman denarii shows its potential for research in numismatics and ancient history. In a manual die study, every coin requires repeated comparisons to all other coins in the collection, or at least to a representative from every already identified die. Samples of sizes comparable to the dataset we considered in this article are thus not feasible for manual die studies, as the number of visual comparison needed to cluster the coins is at best quadratic in their total amount. Using cluster estimations by our method, visual clustering is reduced to corrections and assignments of mislabeled coins. For samples of coins with overall preservation grades comparable to our dataset, this reduces the time investment to a small fraction necessary for a manual die study -- in our experiment by at least $91.8\%$. This makes large die studies feasible, and thus an important historical tool widely available and generally applicable.

\section*{Permissions, data- and code-availability} 
Coin images are reproduced courtesy of auction houses and public collections listed in Table~\ref{table:copyrights}. A catalogue with die analysis for all coins, and links to all images used in this article (published on openly accessible web pages hosted by CoinArchives and public collections), 
as well as code for dissimilarity calculation and clustering 
will be made available upon publication.

\ifCLASSOPTIONcompsoc
  \section*{Acknowledgments}
\else
  \section*{Acknowledgment}
\fi
The authors would like to thank Prof.\ Holly Rushmeier and Prof.\ Robby Tan for helpful discussions.

\ifCLASSOPTIONcaptionsoff
  \newpage
\fi



\bibliographystyle{IEEEtran}
\bibliography{DieStudyIEEE}
%
%
%

%







\newpage

\appendices
\counterwithin{figure}{section}
\counterwithin{table}{section}

\section{Supplementary figures and tables; image copyrights}\label{App}

\begin{table}[!ht]
\caption{Sensitivity to prior hyper-parameters representing cluster size expectation $\mu$ and variance $\nu$ ($\nu=2.4\mu$ except for results in the last column, 500\,000 MCMC iterations, half of them burn-in, thinning to every tenth iteration).
Performance results are stable, increasing $\mu$ will eventually result in overall higher sensitivity and lower FDR (in the sense of Sec.~\ref{Sec:Results}), as estimated total number of clusters will start to decrease.
}
\label{tab:sen}
\centering
\ra{1.2}
\begin{tabular}[h]{@{}lccccc@{}}\toprule 
        & $\mu=4$ & $\mu=5$       & $\mu=6$     & $\mu=7$        & $\mu=6$, $\nu=1.5\mu$\\
\midrule
NMI          & $0.951$ & $0.945$   & $0.957$ & $0.943$    & $0.953$\\
Sens         & $0.789$ & $0.817$   & $0.838$ & $0.855$    & $0.846$\\
FDR          & $0.066$ & $0.149$   & $0.089$ & $0.211$    & $0.135$\\
\bottomrule
\end{tabular}
\end{table}%


\begin{table}[htb]
\caption{Coin images are reproduced with kind permission by the collections, auction houses and photographers listed below. 
Order in figures and subfigures is left-to-right, top-to-bottom.}
\label{table:copyrights}
\centering
\resizebox{\columnwidth}{!}{
\ra{1.2}
\begin{tabular}[h]{@{}llp{9.1cm}@{}}\toprule 
Fig.~\ref{Fig:DifferentOrSame}  
&(a)		&   Tauler \& Fau -  Auction 50, Lot 195\\
&(b)			&   Auktionshaus H. D. Rauch GmbH - E-Auction 23, Lot 2150\\
&(c)		&   Gerhard Hirsch Nachfolger - Auction 298, Lot 546\\
&(d)		&   Bertolami Fine Arts - ACR Auctions - Auction 12, Lot 670\\
\midrule
Fig.~\ref{Fig:CoinPreservation}
&(a)	&	Numismatik Lanz München - Auction 125, Lot 676\\
&(b)	&	Roma Numismatics Ltd - E-Sale 41, Lot 694\\
&(c)	&	Münzkabinett Wien, 785 RÖ 87969\\
\midrule 
Fig.~\ref{Fig:ReverseMotifs} 
&	&	Auktionshaus H. D. Rauch GmbH -  Auction 99, Lot 115\\
&	&	Numismatica Ars Classica - Auction 106, Lot 897\\
&	&	Numismatica Ars Classica -  Auction 62, Lot 2023\\
&	&	Harlan J. Berk, Ltd - Buy or Bid Sale 193, Lot 282\\
&	&	Classical Numismatic Group -  Auction 111, Lot 668\\
&	&	Emporium Hamburg - Auction 79, Lot 255\\
&   &   Nomisma S.p.a. - Auction 47, Lot 136 \\
&	&	Nomisma S.p.a. - Auction 47, Lot 137\\
&	&	American Numismatic Society 1995.11.1577\\
&	&	Numismatik Lanz München - Auction 109, Lot 314\\ 
\midrule
\end{tabular}
}
\end{table}%
\begin{table}[htb]
\centering
\resizebox{\columnwidth}{!}{
\ra{1.2}
\begin{tabular}[h]{@{}llp{9.1cm}@{}}
\midrule 
Fig.~\ref{Fig:SimilarDies}
&&    	Bruun Rasmussen - Auction 852, Lot 5355\\
&&		Nudelman Numismatica - CIP Invest Ltd - Auction 10, Lot 26\\
&&		Nomos -  Auction 2, Lot 173\\
&&		Jesus Vico S.A. - Online Auction 8, Lot 197\\
\midrule
Fig.~\ref{Fig:GPvsORB}/\ref{ImageToLMPrior}
&&	Classical Numismatic Group - Triton XX, Lot 673 \\
\midrule
Fig.~\ref{Fig:matchComparison}
&&	Classical Numismatic Group - Triton XX, Lot 673 \\
&&	Auktionshaus H. D. Rauch GmbH - E-Auction 12, Lot 380\\
&&	Münzkabinett Berlin 18220712\\
&&	Jesus Vico S.A. -  Auction 156, Lot 477\\
\midrule
Fig.~\ref{Fig:PredExpl2}
&(a)&	Gorny \& Mosch Giessener Münzhandlung -  Auction 126, Lot 2275\\
&&		Roma Numismatics Ltd -  E-Sale 66, Lot 882\\
&&		Fritz Rudolf Künker GmbH \& Co. KG - Auction 326, Lot 1329\\
&&		Jesus Vico S.A. -  Online Auction 5, Lot 92\\
&&		Jean Elsen \& ses Fils S.A. -  Auction 103, Lot 144\\
&(b)&	Münzkabinett Berlin 18220746 \\
&&  	Noble Numismatics Pty Ltd -  Auction 116, Lot 4544 \\
&(c)&	Auktionshaus Münzhandlung Sonntag - Auction 3, Lot 118\\	
&&		Classical Numismatic Group - Electronic Auction 457, Lot 255\\
&&		Naville Numismatics Ltd - Auction 58, Lot 511\\
&&		Gorny \& Mosch -  Auction 147, Lot 1996\\
&&		Solidus Numismatik -  Online Auction 7, Lot 223\\
&&		Numismatik Naumann -  Auction 53, Lot 698\\
&&		Gorny \& Mosch - Auction 208, Lot 1992\\
&&		Auktionshaus H. D. Rauch GmbH -Auction 90, Lot 351\\
&&		Numismatik Naumann - Auction 73, Lot 463\\
\midrule
Fig.~\ref{Fig:PredExplMixes}
&(a)&      Gorny \& Mosch Giessener Münzhandlung - Auction 152, Lot 2034\\
&& 		Numismatik Naumann -  Auction 70, Lot 390\\
&&		Heidelberger Münzhandlung Herbert Grün e.K.  - Auction 66, Lot 109\\
&&  	Auktionshaus Gärtner GmbH \& Co.\ KG - Auction 32, Lot 34451\\
&&		Áureo \& Calicó - Auction 247, Lot 1026\\
&&		Gerhard Hirsch Nachfolger - Auction 271, Lot 2249\\
&&		Roma Numismatics Ltd - E-Sale 46, Lot 550\\
&&		Inasta - Auction 11, Lot 291\\
&& 		Stephen Album Rare Coins - Auction 35, Lot 511\\
&&		Inasta - Auction 58, Lot 684\\
&&		Inasta - Auction 12, Lot 711\\
&(c)&   Leu Numismatik AG - Auction 5, Lot 336\\
&&    Auktionshaus H. D. Rauch GmbH -  Auction 99, Lot 115\\
&&    Numismatica Ars Classica -  Auction 52, Lot 349\\
&&    Münzkabinett Berlin 18220747\\
&(d)&    Münzsammlung, Albert-Ludwigs-Universität, Freiburg 00704\\
&&     Classical Numismatic Group - Electronic Auction 105, Lot 150\\
&&     Classical Numismatic Group - Electronic Auction 182, Lot 213\\
&&     Auktionshaus H. D. Rauch GmbH - E-Auction 20, Lot 261\\
&(e)&    Savoca Numismatik GmbH \& Co. KG - Special Auction 79, Lot 507\\
&&    Auktionshaus H. D. Rauch GmbH - Auction 77, Lot 378\\
&&    Classical Numismatic Group - Electronic Auction 459, Lot 459\\
&(f)&   Münzkabinett Berlin 18220718\\
&&    Auktionshaus H. D. Rauch GmbH - Summer Auction 2007, Lot 375\\
&&    Classical Numismatic Group - Electronic Auction 313, Lot 216\\
&&    Classical Numismatic Group - Auction 105, Lot 826\\
&(g)&	Aurea Numismatika - Auction 13, Lot 319\\
&&		Gorny \& Mosch Giessener Münzhandlung - Auction 165, Lot 1857\\
&&		Roma Numismatics Ltd - E-Sale 26, Lot 655\\
&&		Classical Numismatic Group -  Auction 88, Lot 1244\\
&&		Numismatica Ars Classica -  Auction 46, Lot 997\\
&&		Gorny \& Mosch - Auction 118, Lot 2059\\
&(h)&    Roma Numismatics Ltd -  E-Sale 61, Lot 643\\
&&    Auktionen Münzhandlung Sonntag  - Auction 28, Lot 20\\
&&    Noble Numismatics Pty Ltd  - Auction 123, Lot 3339\\
&&    Savoca Numismatik GmbH \& Co. KG - Auction 16, Lot 435\\
&&    Auktionshaus H. D. Rauch GmbH - E-Auction 20, Lot 263\\
&&    Classical Numismatic Group - Auction 102, Lot 875\\
&&    Tauler \& Fau - Auction 55, Lot 5323\\
&&    Münzkabinett Berlin 18220715\\
&(i)&     Gorny \& Mosch Giessener Münzhandlung -  Auction 108, Lot 1714\\
\midrule
Fig.~\ref{Fig:badcoins}
&&		American Numismatic Society 1956.127.3\\
&&		Classical Numismatic Group - Electronic Auction 327, Lot 908\\
&&		Dix Noonan Webb Ltd - 22 Oct 2009 Auction - Sale DC1, Lot 59\\
&&		Classical Numismatic Group - Electronic Auction 390, Lot 476\\
&&      Jean Elsen \& ses Fils S.A. - Auction 95, Lot 366\\
&&		Tauler \& Fau - Auction 2, Lot 239\\
&&		Auktionshaus Felzmann - Auction 163, Lot 20155\\
&&		Bruun Rasmussen - Online Auction 2027, Lot 5033\\
\midrule
Fig.~\ref{Fig:Identicals} 
&(a)&	Classical Numismatic Group - Triton VI, Lot 822\\
&(b)&	Roma Numismatics Ltd -  Auction XVI, Lot 679\\
&(c)&	Roma Numismatics Ltd - Auction XVIII, Lot 1100\\
\midrule
Fig.~\ref{Fig:Reverses}
&(a)&	Dr. Busso Peus Nachfolger - Auction 378, Lot 454\\
&(b)&	Harlan J. Berk, Ltd - Buy or Bid Sale 206, Lot 201\\
&(c)&	Gorny \& Mosch Giessener Münzhandlung - Auction 180, Lot 367\\
\bottomrule
\end{tabular}
}
\end{table}%

\end{document}